%% file: DRAFT/Draft_CR.tex
\documentclass[sigconf]{acmart}
\AtBeginDocument{%
  }

\newcommand{\highlight}[2]{\colorbox{#1!20}{\textbf{#2}}}
\usepackage{url}
\usepackage{makecell}
\usepackage{algpseudocode}
\usepackage{algorithm}
\usepackage{adjustbox, array}
\usepackage{amsfonts}
\usepackage{graphicx}      
\usepackage{amsmath}       
\usepackage{adjustbox}     
\usepackage{enumitem}
\usepackage{booktabs}
\usepackage{multirow}
\usepackage{pifont}
\usepackage{balance}
\newcommand{\cmark}{\ding{51}}%
\newcommand{\xmark}{\ding{55}}%

\usepackage{mathtools}
\mathtoolsset{showonlyrefs}

\def\ACRONYM{SLGNN}

\usepackage[colorinlistoftodos]{todonotes}

\copyrightyear{2026}
\acmYear{2026}
\setcopyright{cc}
\setcctype{by}
\acmConference[WWW '26]{Proceedings of the ACM Web Conference 2026}{April 13--17, 2026}{Dubai, United Arab Emirates}
\acmBooktitle{Proceedings of the ACM Web Conference 2026 (WWW '26), April 13--17, 2026, Dubai, United Arab Emirates}
\acmPrice{}
\acmDOI{10.1145/3774904.3792675}
\acmISBN{979-8-4007-2307-0/2026/04}

\begin{document}

\title{Stuart-Landau Oscillatory Graph Neural Network}



\author{Kaicheng Zhang}
\email{K.Zhang-60@sms.ed.ac.uk}
\affiliation{%
  \institution{School of Mathematics and Maxwell Institute\\ University of Edinburgh}
  \city{Edinburgh}
  \country{UK}
}

\author{David N. Reynolds}
\email{d.reynolds@uw.edu.pl}
\affiliation{%
  \institution{Institute of Applied Mathematics and Mechanics\\University of Warsaw}
  \city{Warsaw}
  \country{Poland}
}

\author{Piero Deidda}
\email{piero.deidda@sns.it}
\affiliation{%
  \institution{Gran Sasso Science Institute}
  \city{L’Aquila}
  \country{Italy}
  }
  \affiliation{%
  \institution{Scuola Normale Superiore}
  \city{Pisa}
  \country{Italy}
  }

\author{Francesco Tudisco}
\email{f.tudisco@ed.ac.uk}
\affiliation{\institution{School of Mathematics and Maxwell Institute, \\University of Edinburgh,}
  \city{Edinburgh}
  \country{UK}}
\affiliation{\institution{Miniml.AI}
  \city{Edinburgh}
  \country{UK}}

\renewcommand{\shortauthors}{Kaicheng Zhang, David N. Reynolds, Piero Deidda, and Francesco Tudisco}





\newpage

\begin{abstract}
Oscillatory Graph Neural Networks (OGNNs) are an emerging class of physics-inspired architectures designed to mitigate oversmoothing and vanishing gradient problems in deep GNNs. In this work, we introduce the Complex-Valued Stuart-Landau Graph Neural Network (\ACRONYM), a novel architecture grounded in Stuart-Landau oscillator dynamics. Stuart-Landau oscillators are canonical models of limit-cycle behavior near Hopf bifurcations, which are fundamental to synchronization theory and are widely used in e.g.\ neuroscience for mesoscopic brain modeling. Unlike harmonic oscillators and phase-only Kuramoto models, Stuart-Landau oscillators retain both amplitude and phase dynamics, enabling rich phenomena such as amplitude regulation and multistable synchronization. The proposed \ACRONYM{} generalizes existing phase-centric Kuramoto-based OGNNs by allowing node feature amplitudes to evolve dynamically according to Stuart-Landau dynamics, with explicit tunable hyperparameters (such as the Hopf-parameter and the coupling strength) providing additional control over the interplay between feature amplitudes and network structure. We conduct extensive experiments across node classification, graph classification, and graph regression tasks, demonstrating that \ACRONYM{} outperforms existing OGNNs and establishes a novel, expressive, and theoretically grounded framework for deep oscillatory architectures on graphs. \\
The code and hyperparameters for \ACRONYM{} are available at \url{https://github.com/kevvzhang/StuartLandauGNN}
\end{abstract}

\maketitle

\section{Introduction}
Graph neural networks (GNNs) are a class of neural networks designed to learn from graph-structured data by modeling the mutual influence between each node and its neighbors. These architectures have demonstrated effectiveness across diverse domains, ranging from social network analysis \cite{fanGraphNeuralNetworks2019} to molecular property prediction \cite{pmlr-v70-gilmer17a} and recommendation systems \cite{pengSVDGCNSimplifiedGraph2022,damianouGraphFoundationModels2024}.

However, classical GNN architectures, such as graph convolutional networks (GCN) \cite{kipfSemiSupervisedClassificationGraph2016} and graph attention networks (GAT) \cite{velickovicGraphAttentionNetworks2017}, suffer from multiple pathologies stemming from their message-passing paradigm for aggregating features from neighboring nodes. These limitations include oversmoothing \cite{oonoGraphNeuralNetworks2019, ntRevisitingGraphNeural2019, ruschSurveyOversmoothingGraph2023, zhang2025rethinking}, overcorrelation \cite{rothRankCollapseCauses2023}, oversquashing \cite{giovanni2024how, nguyenRevisitingOversmoothingOversquashing2022}, and vanishing gradients \cite{ruschGraphCoupledOscillatorNetworks2022}.

To address these challenges, numerous alternative architectures have been proposed, such as Graph Neural ODEs that inherit properties from the physical systems on which they are based \cite{poli2019graph,xhonneuxContinuousGraphNeural2019, chamberlainGRANDGraphNeural2021, Eliasof, thorpeGrandGraphNeural2022}. Among them, a particularly promising class is the oscillatory graph neural networks (OGNN), based on coupled oscillator systems \cite{nguyenCoupledOscillatorsGraph2023, ruschGraphCoupledOscillatorNetworks2022}. In these models, each node feature is represented as an oscillator with coupling factors determined by the graph structure, drawing inspiration from models of electrical activity in biological neural networks.
These architectures have been shown to mitigate several pathologies, including vanishing and exploding gradients as well as oversmoothing \cite{ruschGraphCoupledOscillatorNetworks2022, nguyenCoupledOscillatorsGraph2023}.

In this work, we introduce the Stuart-Landau Graph Neural Network (\ACRONYM), a novel architecture based on complex-valued coupled Stuart-Landau oscillator dynamics. Stuart-Landau oscillators are canonical models of limit-cycle behavior near Hopf bifurcations, generalizing the well-known Kuramoto model of coupled oscillators. Critically, while the Kuramoto model evolves only phase (frequency) dynamics, the Stuart-Landau framework captures both amplitude and phase evolution. This fundamental difference gives rise to substantially richer dynamics: while Kuramoto models converge only to frequency-synchronized equilibrium states, Stuart-Landau systems exhibit a diverse repertoire of stable states---including frequency-synchronized and asynchronous configurations, amplitude-death and active states---whose emergence depends on the interplay between system hyperparameters. This increased expressiveness, rooted in the model's ability to capture the nonlinear amplitude-stabilization mechanisms observed in biological neural oscillators, translates into enhanced learning capacity for \ACRONYM{} compared to existing OGNN models.

The contributions of this paper include the following:
\begin{itemize}[topsep=0pt, leftmargin=*,itemsep=0pt]
    \item We provide a comprehensive overview of three oscillator types (Stuart-Landau, Kuramoto, and harmonic), detailing their dynamical properties and the implications on the learning capabilities.
    \item We introduce \ACRONYM, a new OGNN framework that models both amplitude and phase dynamics through Stuart-Landau coupled oscillators, enabling richer behaviors beyond the Kuramoto- or harmonic-based OGNNs.
    \item We evaluate the architecture using a bespoke IMEX time-stepping scheme that achieves 100$\times$ speedup over adaptive solvers while maintaining stability and performance.
    \item Through extensive experiments across node classification, graph classification, and graph regression tasks, we demonstrate that \ACRONYM{} achieves state-of-the-art performance by effectively leveraging deeper architectures. 
\end{itemize}

\section{Related Work}

Continuous-depth GNN models based on graph ordinary differential equations (ODEs) have demonstrated significant success in circumventing known GNN limitations. Standard architectures like GATs can be interpreted as explicit discretizations of diffusion equations on graphs, and the corresponding continuous-time Graph Neural Diffusion (GRAND) model \cite{chamberlainGRANDGraphNeural2021} has been shown to suffer substantially less from pathologies such as oversmoothing observed in discrete counterparts. Further improvements arise from augmenting the diffusion dynamics with reaction and advection terms, which can be rigorously proven to prevent convergence to node-wise constant steady states \cite{thorpeGrandGraphNeural2022,GREAD_Choi, eliasof2024featuretransportation, GraphNeuralReactionDiffusionModels_Eliasof}. Indeed, from the perspective of a dynamical system,
the increasing similarity of the features in the long-time asymptotic is often a synonym of convergence of the trajectories of the different nodes toward a single steady state. This situation is directly linked to oversmoothing and always occurs when considering diffusion systems on a connected graph, in which case the unique steady state is the uniform one with magnitude depending on the initial condition. However, if the diffusion operator is learned in such a way as to almost disconnect the graph according to the task at hand, then convergence to the stable equilibrium can be significantly slowed. Additionally, adding reaction or advection terms to the model makes the steady state of the system dependent on the initial one, thus avoiding phenomena like oversmoothing and overcorrelation.

A particularly promising class of continuous Neural ODEs consists of physics-inspired oscillatory models.
Coupled Oscillatory Recurrent Neural Network (coRNN) \cite{ruschCoupledOscillatoryRecurrent2020} employs coupled damped second-order harmonic oscillators to mitigate the vanishing and exploding gradient problems in residual neural networks for tackling long-sequence tasks. Unified Invertible Coupled Oscillatory RNN (UnICORNN) \cite{ruschUnICORNNRecurrentModel2021} generalizes coRNN as a Hamiltonian system, improving memory efficiency while ensuring invertibility in time. Building on this, Linear Oscillatory State-Space models (LinOSS) \cite{ruschOscillatoryStateSpaceModels2024} translates these dynamics from the standard RNN setting into a linear state-space model, enabling highly efficient computation via parallel scans rather than step-by-step recurrence.
\citeauthor{lanthalerNeuralOscillatorsAre2023} \cite{lanthalerNeuralOscillatorsAre2023} further demonstrates the universality of the harmonic oscillatory networks for approximating underlying functions. Apart from harmonic-oscillator-based approaches, Artificial Kuramoto Oscillatory Neurons (AKOrN) \cite{miyatoArtificialKuramotoOscillatory2024} propose a general-purpose oscillatory neuron as unit vectors evolving according to Kuramoto dynamics and with great adaptability with existing layer architectures such as convolutions and attention mechanisms.

For the application of oscillatory models to graph tasks, the Graph Coupled Oscillator Network (GraphCON) \cite{ruschGraphCoupledOscillatorNetworks2022} uses damped harmonic oscillators similar to those of coRNN, while Kuramoto Graph Neural Networks (KuramotoGNN) \cite{nguyenCoupledOscillatorsGraph2023} model each feature as a fixed-amplitude oscillator with nonlinear coupling between oscillators. Both approaches provably reduce or eliminate oversmoothing as well as exploding and vanishing gradient problems. Notably, in coupled oscillator systems, a steady state is always the one where both phases and amplitudes are synchronized and possibly zero if damping factors are included. Therefore, in order to avoid oversmoothing/overcorrelation problems, oscillatory models admitting steady states other than the fully synchronized one are desirable. 
In this setting, our proposed Stuart-Landau architecture is of particular interest. Indeed, by enabling dynamic amplitude evolution alongside phase dynamics, it enriches the space of possible equilibrium states and enhances model expressiveness.

\begin{table*}[t]
\centering
\setlength{\tabcolsep}{1mm}
\input{Tables/oscillator_property_comparison_v3}
\caption{Dynamics that can be modeled by different oscillators and the non-oscillatory baseline. The magnitude behaviour and convergence figures show the dynamics of the oscillator on a complex plane. The black dot indicates the origin, and the dashed lines indicate some magnitude equilibrium position. The phase behaviour figures show the evolution of the phase over time.}
\label{tab:oscillator_comparison}
\end{table*}

\section{Three types of neural oscillators}
In this section, we present the Stuart-Landau oscillator dynamical system and provide an overview of two other prominent types of oscillators used for modeling physics-inspired graph neural networks: the Kuramoto oscillator and the (un)damped harmonic oscillator. We discuss their main properties and compare their relative benefits and deficits.
\subsection{The Stuart-Landau Oscillator}
A Stuart-Landau oscillator describes the behavior of a limit-cycle oscillator near a Hopf bifurcation. The equation is given by
\begin{align}
    \dot z(t)&=(\alpha+i\omega-(\beta+i\gamma)|z(t)|^2)z(t),\label{SLosc1}\\
    &\alpha,\omega,\beta,\gamma \in \mathbb{R}, \ \ z(t)\in \mathbb{C}.\nonumber
\end{align}
A Hopf bifurcation occurs when at a phase-transition a fixed point gives rise to a stable limit cycle. To see how each of the four parameters $(\alpha,\omega,\beta,\gamma)$ play a role in the dynamics and asymptotic outcomes let us break down the oscillator into its amplitude and phase values $z(t)=r(t)e^{i\phi(t)} \in \mathbb{C}$. The amplitude equation is given~by
\begin{align}\label{amp}
    \dot r(t)=(\alpha-\beta r^2(t))r(t).
\end{align}
Only two parameters play a role in the amplitude behavior, which gives us 4 main cases, \cite{MPRT-SL,panteleyStabilityRobustnessStuartlandau2015,Reynolds_2025}: 
\begin{enumerate}[topsep=0pt, leftmargin=*,itemsep=0pt]
    \item $(\alpha>0,\beta>0)$: There is a fixed point of \eqref{amp} at $r_{\infty}=\sqrt{\alpha/\beta}$ which is stable and $r_{\infty}=0$ which is unstable.
    \item $(\alpha< 0, \beta\geq 0)$: There is a stable fixed point at $r_{\infty}=0$ for which all $r(t)$ converge to exponentially fast.
    \item $(\alpha>0, \beta\leq 0)$: There is a fixed point at $r_{\infty}=0$ which is unstable, and if $r_0>0$, then the amplitude experiences unbounded growth, $r(t)\to \infty$.
    \item $(\alpha<0, \beta<0)$: The fixed point $r_{\infty}=0$ is still stable, but the basin of attraction is confined to $r_0^2<\alpha/\beta$, while $r_{\infty}^2=\alpha/\beta$ is an unstable fixed point where if $r_0^2>\alpha/\beta$, then $r(t) \to \infty$.
\end{enumerate}
The standard form for a Stuart-Landau oscillator assumes $\beta>0$ to avoid situations in which the amplitude undergoes unbounded growth. This leads us to focus on cases (1) and (2). Case (1) can be identified with the Orbit and Amplitude Growth columns of Table \ref{tab:oscillator_comparison}, with magnitude convergence occurring at an exponential rate. Case (2) leads to Amplitude Death, again at an exponential rate. However, in the above four cases we left off the case of $\alpha=0$ as, $\alpha$, known as the Hopf-parameter has its critical value at zero. The critical case gives
\begin{align}
    \dot r(t)=-\beta r^3(t)\label{crit}
\end{align}
For $\beta>0$, solving \eqref{crit} gives
\begin{align}
    r(t)= (2\beta t+c)^{-1/2},
\end{align}
we don't see exponential convergence, but rather convergence to zero at an algebraic rate as in Table \ref{tab:oscillator_comparison}.

Now let us analyze the phase behavior,
\begin{align}\label{phase}
    \dot \phi(t)=\omega-\gamma r^2.
\end{align}
Again, the phase equation only depends on two parameters, however the dependence on the amplitude also plays a role. The parameter $\omega$ is known as the natural frequency of the oscillator and in the subcritical regime $(\alpha<0, \beta>0)$ since $r(t) \to 0$ exponentially fast the resulting phase dynamics lead to $\dot \phi(t) \to \omega$ also exponentially fast giving an almost constant phase speed as the oscillator converges to zero. The parameter $\gamma$ is known as the phase shift parameter and when $\gamma=0$ we see that $\dot \phi=\omega$ remains fixed, while for $\gamma>0$ the oscillator is slowed down and respectively sped up for $\gamma<0$. Now, if we are in the supercritical regime $(\alpha>0, \beta>0)$ we have $\dot \phi \to \omega-\gamma\frac{\alpha}{\beta}$ so that $\gamma$ provides for a phase shift reliant also on the amplitude parameters $\alpha$ and $\beta$.

Again, focusing on the critical case of $\alpha=0, \beta>0$ we see that $r^2(t)=(2\beta t+c)^{-1}$ and hence
\begin{align}\label{alge}
    \dot \phi(t)=\omega-\frac{\gamma}{2\beta t+c},
\end{align}
which again implies that the phase settles to the natural frequency $\omega$, as $t \to \infty$, but at an algebraic rate. 
\subsubsection{Coupled Stuart-Landau Oscillators}
Coupling identical oscillators of the form \eqref{SLosc1} gives the following equation
\begin{align}\label{eq:SL1}
    \dot{z}_j = (\alpha + i\omega - (\beta+i\gamma)|z_j|^2)z_j +  \sum_{l=1}^N A_{lj}(z_l - z_j), 
\end{align}
for $j = 1, \dots, N$, where $A$ represents the connectivity matrix. Separating into their respective amplitude and phase equations gives
\begin{align}
&\dot r_j=(\alpha-\beta r_j^2)r_j+\sum_{l=1}^NA_{lj}(\cos(\phi_l-\phi_j)r_l-r_j),\label{eq:SLr1}\\
    &\dot \phi_j=\omega-\gamma r_j^2+\sum_{l=1}^NA_{lj}\frac{r_l}{r_j}\sin(\phi_l-\phi_j).\label{eq:SLphi1}
\end{align}
When the system is decoupled and each of the $N$ oscillators converge to their limit-cycle (supercritical case) or to the zero fixed point (subcritical case). In the supercritical case, the initial data of each oscillator will determine where on the limit-cycle each oscillator can be found. Once the coupling is activated, again in the supercritical case, if the connectivity matrix is connected, one can observe that 
\begin{align}
&z_j(t)=z_k(t) \ \ \text{for all} \ j,k=1,\ldots, N,\\
&r_{j,\infty} =\sqrt{\alpha/\beta} \quad \dot\phi_j=\omega-\gamma\sqrt{\alpha/\beta}
\end{align}
represents a stable synchronized asymptotic state for the system, \cite{MPRT-SL,Reynolds_2025}, represented in the final column of Table \ref{tab:oscillator_comparison}. However, depending on the connectivity matrix, other stable asymptotic states representative of incoherence (uniformly distributed around a smaller limit cycle) and other symmetric configurations can coexist with the fully synchronized state, \cite{arenas-diaz-kurths-moreno-zhou}, represented in the second to last column of Table \ref{tab:oscillator_comparison}.

On the other hand, in the subcritical case, the amplitudes converge to zero exponentially fast, regardless of the coupling, while the phase behavior can be characterized by \eqref{eq:SLphi1} where each $\frac{r_l}{r_j}$ has converged to a finite value and thus can be absorbed into the network structure and each $r_j \to 0$ exponentially fast will no longer play a role in the phase dynamics \cite{MPRT-SL}. Thus one can see that the fully synchronized state $\dot\phi_j=\omega$ can be recovered in this regime, as well as the incoherent and other symmetric states found in \cite{doi:10.1137/23M155400X}.

Now, focusing on the critical case. For the fully synchronized state, it is easy to see that if all $\phi_j=\phi_k$, and all $r_j=r_k$, then the respective dynamics are reduced to
\begin{align}
    \dot r_j=(\alpha-\beta r_j^2)r_j, \quad \text{and} \quad 
    \dot \phi_j=\omega-\gamma r_j^2,
\end{align}
for each $j=1,\ldots,N$ and hence the critical value $\alpha=0$ yields an algebraic speed of relaxation as in \eqref{alge}. 

\subsection{Kuramoto oscillators}
Careful observation of \eqref{eq:SLphi1} would lead the reader to notice the similarity to the famous Kuramoto model of phase synchronization. In fact, Kuramoto derived his model from the system \eqref{eq:SL1} via a phase reduction that removed the amplitude dependence of the oscillators \cite{Kuramoto:1975ebm}. The reduction, which is valid in the week coupling regime, involves taking the limits $\alpha,\beta\to \infty$ while keeping the ratio $\frac{\alpha}{\beta}$ fixed, forces each oscillator to take the amplitude $r_j=\sqrt{\alpha/\beta}$ so that the only relevant aspect of the model is the phase behavior. This is known as a phase reduction and gives rise to the model for identical Kuramoto oscillators (KO)
\begin{align}\label{KM}
    \dot \phi_j=\omega+\sum_{l=1}^NA_{lj}\sin(\phi_l-\phi_j).
\end{align}
We begin here with the coupled form as for an individual KO it is clear that $\dot \phi=\omega$ is the phase behavior, and there is no amplitude behavior, thus it always evolves on an orbit. From a mathematical perspective, this phase reduction greatly simplifies the model and allows for a more complete analysis of the dynamics. Under a symmetric graph structure, the model obeys a gradient flow structure given by
\begin{align}
    \dot \Phi&=-\nabla E(\Phi),\\
    E(\Phi)&=\sum_{j\neq k=1}^N \left(\frac{1}{2}A_{jk}\cos(\phi_j-\phi_k)\right)-\sum_{j=1}^N\omega\phi_j
\end{align}
 The fully synchronized state represents the global minimum of the energy of the gradient flow and thus is always a steady state of the system and is exponentially stable. However, depending on the graph structure, other frequency synchronized states such that $\dot\phi_j=\dot\phi_k$ for all $j,k=1,\ldots,N$ while $\phi_j \neq \phi_k$ in general can also exist and be exponentially stable, e.g. for a ring structure \cite{arenas-diaz-kurths-moreno-zhou}. In \cite{Taylor_2012} it was first shown that if a graph is dense enough, then the only stable outcome is the fully synchronized state, which is thus a global attractor, while in \cite{doi:10.1137/23M155400X} it was shown that for a particular balanced graph structure, there are infinitely many distinct exponentially stable equilibria. Even further, under other specific graph structures there exist so-called twisted states and other balanced states which are not Lyapunov stable due to the existence of multiple zero eigenvalues, but nonetheless can be shown to be nonlinearly stable or neutrally stable \cite{Sclosa_2024,10.1063/1.2165594}.

\subsection{The Harmonic Oscillator}
Our third type of oscillator is the Harmonic Oscillator (HO) which is given by a linear second order real ordinary differential equation
\begin{align}\label{Harmonic}
    \ddot x(t)+2\zeta \omega_0\dot x(t)+\omega_0^2x(t)=0
\end{align}
The value $\zeta\geq 0$ determines the damping of the system and yields 4 regimes.
\begin{enumerate}[topsep=0pt, leftmargin=*,itemsep=0pt]
    \item (Undamped $\zeta=0$): The solution is given by $x(t)=A\sin(\omega_0t+\phi)$, where the amplitude $A>0$ and phase $\phi$ are prescribed by the initial data.
    \item (Underdamped $\zeta \in (0,1)$: The solution is given by $x(t)=Ae^{-\zeta\omega_0t}\sin(\omega_dt+\phi)$ where $\omega_d=\omega_0\sqrt{1-\zeta^2}$ slows the oscillation speed compared to the undamped case. The solution eventually stops as $x(t)\to 0$ exponentially fast at rate $\zeta\omega_0$.
    \item (Critically damped $\zeta=1$): The solution is given by $x(t)=(c_1+c_2t)e^{-\omega_0t}$ as $\omega_0$ is a repeated root of the characteristic equation and solutions converge to zero exponentially fast without oscillations.
    \item (Overdamped $\zeta>1$): The solution is given by $x(t)=c_1e^{r_1t}+c_2e^{r_2t}$ where $r_1$ and $r_2$ are negative real roots to the characteristic equation which drive the solution directly to zero with no oscillations.
\end{enumerate}
As oscillations are the main object of concern, we do not further consider the critically damped or overdamped cases since they undergo exponential decay to zero without oscillation. The undamped case retains the first two regimes where $\zeta\in [0,1)$ will be the focus. Case (1) can be identified with the second column of Table \ref{tab:oscillator_comparison} as the lack of damping retains a constant amplitude. Case (2) on the otherhand gives exponential decay to zero and thus the Amplitude Death and Exponential convergence columns of Table \ref{tab:oscillator_comparison} are satisfied.

\subsubsection{Coupled Harmonic Oscillators}
Coupling identical oscillators of the form \eqref{Harmonic} gives the following equation
\begin{align}
    \ddot x_j+c\dot x_j+\sum_{l=1}^NA_{lj}(x_j-x_l)=0.
\end{align}
For symmetric coupling $A$, it is classical, \cite{Rao}, to show that $x(t)=(x_1(t), \ldots,x_n(t))$ is given by
\begin{align}
    &x(t)=\sum_{i=1}^N q_i(t)\, v_i, \qquad q_i(t) = v_i^\top x(t),\\
    &q_i(t)=e^{-\zeta_i\omega_i t}\left( q_i(0)\cos(\omega_{d,i}t)
+ \frac{\dot q_i(0)+\zeta_i\omega_i q_i(0)}{\omega_{d,i}}\sin(\omega_{d,i}t)\right),
\end{align}
where $(\omega_i^2,v_i)$ are the respective eigenvalue-eigenvector pair of $A$, and $\zeta_i=\frac{c}{2\omega_i}$ and again $\omega_{d,i}=\omega_i\sqrt{1-\zeta_i^2}$. In this way it is easy to see that in the undamped case $c=0$, the solution is given by $x(t)=\sum_i (A_i\cos(\omega_it)+B_i\sin(\omega_it))v_i$ with $A_i$ and $B_i$ completely determined by the initial data. On the other hand, in the damped case we see exponential decay to zero. Therefore both incoherence and synchronization are possible outcomes.

\subsection{Phenomenological comparison of learning dynamics for the three oscillators}
\subsubsection{Individual oscillators:} \textit{Active regimes:} From a qualitative perspective the undamped regime $(\zeta=0)$ of the HO is most similar to the supercritical regime $(\alpha>0)$ of an SL oscillator as well as the behavior of a KO. Indeed, An individual KO oscillator and undamped HO have a constant phase velocity and no amplitude dynamics. If the initial data for an SL oscillator is not found on the stable limit-cycle, then the nonlinear damping term $-\beta|z(t)|^2z(t)$ creates a correction mechanism for the amplitude behavior and a phase-drift that relaxes to the resulting limit-cycle determined by the system parameters $(\alpha,\beta,\omega,\gamma)$. In comparison, the KO has no amplitude behavior due to it being a purely phase-oscillator, while the amplitude of an HO is given by its initial data. This means that the SL oscillator is robust with respect to perturbations, while the HO yields periodic behavior once again, but at a new amplitude. The phase behavior of both the HO and SL oscillator are determined by the parameters however if the SL phase drift parameter $\gamma$ is nonzero, then the final phase speed is affected by the amplitude behavior in contrast to the HO which has constant amplitude.

\textit{Decay regimes:} A single underdamped HO with $\zeta \in (0,1)$ undergoes oscillations, but exponentially decays to zero. The same is true for a subcritical $(\alpha<0)$ SL oscillator. However, the SL oscillator at critical value $\alpha=0$ decays to zero at an algebraic rate. The KO is restricted to the torus and does not have amplitude dynamics.

\subsubsection{The coupled systems:} \textit{Active regimes:} Starting with coupled undamped HOs, we see that the phase dynamics of each oscillator are entirely determined by the eigenvalues of the connectivity matrix $A$ and that the amplitudes are still determined by the initial data. Further, as seen in \cite{ruschGraphCoupledOscillatorNetworks2022}, the energy of the system is preserved. Therefore synchronization only occurs if the connectivity matrix has all identical eigenvalues, and each node has identical initial data, otherwise the behavior will in general be quasiperiodic, and nodes should remain distinguishable. 

Coupled Kuramoto oscillators have been seen to have a wealth of stable equilibria which are not synchronized states when the connectivity matrix is not too dense. However, the synchronized state always exists and is always exponentially stable with a large basin of attraction. Therefore as long as initial data is not found in the basin of attraction of the synchronized state the nodes should remain distinguishable, but there is a possibility for oversmoothing if perturbations lead to falling into the basin of attraction of the synchronized state.

As the Kuramoto model is a reduction of the Stuart-Landau system, we see that each of the stable equilibria of the KO model are also stable equilibria of the SL system, however, due to the evolution of SL in the complex plane rather than on a fixed torus there is not only the possibility of other fixed points, but a greater ability to explore the latent space before settling to the equilibria.

\textit{Decay regimes:} As the Kuramoto model has no amplitude death, we only compare the underdamped HO $(c \in (0,1)$ and (sub)critical SL oscillators $(\alpha \leq 0)$ here. In these regimes we see both the SL oscillators and HO converge to zero. For all $c \in (0,1)$ we see that oscillations continue, but each oscillator decays to zero at rate $\zeta_i=\frac{c}{2\omega_i}$ and hence one expects oversmoothing to occur. The same is true for subcirtical $(\alpha<0)$ SL oscillators which decay to zero at an exponential rate. However, only the SL oscillatory system allows for even the synchronized state to avoid oversmoothing. In the critical case $\alpha=0$, the amplitude dynamics converge to zero at an algebraic rate rather than exponential.

\section{Start Landau Graph Neural Network}
The Stuart-Landau Graph Neural Network (\ACRONYM{}) adapts the continuous-time dynamics of coupled Stuart-Landau oscillators for graph-based learning tasks. To achieve this, we formulate the node feature evolution as a graph ODE:
\begin{align} \label{eq:slgnn_continuous}
    \dot z(t)&=(\alpha+i\omega-(\beta+i\gamma)|z(t)|^2)z(t) + F_\theta(z(t),t), 
\end{align}
where $z(0)$ is an initial feature matrix produced by a complex-valued MLP encoder.
The first term governs the local Stuart-Landau dynamics for each node, while $F_\theta(z(t),t)$ represents a learnable, time-dependent nonlinear coupling between neighbouring nodes. However, a significant challenge in implementing this model is the numerical instability introduced by the cubic term $|z(t)|^2z(t)$. 
Standard explicit numerical solvers, such as the Euler method, requires very small time steps to maintain stability, leading to model underperformance. While adaptive time-step solvers, such as the Dormand-Prince method, can handle such stiffness, they are often very slow and susceptible to step size underflow during training.

To overcome these issues, we propose a bespoke implicit-explicit (IMEX) numerical scheme. The core idea of an IMEX scheme is to treat the stiff parts of the ODE implicitly to ensure stability, while treating the non-stiff parts explicitly to maintain computational efficiency. A naive implicit approach might use Newton's method to solve the entire local oscillatory term
\begin{align}\label{eq:slgnn_discrete}
Z^{(l+1)} &= Z^{(l)} + dt (\alpha+i\omega-(\beta+i\gamma)|Z^{(l+1)}|^2)Z^{(l+1)},
\end{align}
while treating the coupling term explicitly. $Z^{(l)}$ denotes the complex-valued feature matrix at the $l$-th layer. Although feasible, this approach necessitates the computation of a 2D Jacobian and takes more iterations to converge. Therefore, we further refine the IMEX scheme by decomposing Equation \eqref{eq:slgnn_discrete} into their magnitude and phase terms, and only implicitly solve the stiff cubic magnitude term. This further reduces the dimensionality of the implicit part to one, as the phase update can be kept explicit. We present a Pseudocode of \ACRONYM{} in Algorithm 1.

\begin{algorithm}[h]\label{alg:slgnn_full_model}
\begin{flushleft}
\scriptsize
\ttfamily
\textcolor{teal}{\# $X^{(0)}$: initial real-valued node features}\\
\textcolor{teal}{\# ENC: complex-valued MLP encoder; DEC: real-valued MLP decoder} \\
\textcolor{teal}{\# $F_\theta^{(l)}$: coupling function; $L$: number of layers} \\
\textcolor{teal}{\# $dt$: time step; $\alpha, \beta, \omega, \gamma$: oscillator parameters} \\
\vspace{2mm}

\textcolor{teal}{\# Encode initial features into the complex domain}\\
$Z^{(0)} \leftarrow \text{ENC}(X^{(0)})$\\
\vspace{2mm}

\textcolor{teal}{\# Iterate through L layers using the IMEX time-stepping scheme}\\
\textbf{for} $l = 0$ \textbf{to} $L-1$ \textbf{do}\\
    \hspace*{4.3mm}\textcolor{teal}{\# Explicit step for the non-stiff coupling term}\\
    \hspace*{4.3mm}$\tilde Z^{(l)} \leftarrow Z^{(l)} + dt ~ F^{(l)}_\theta(Z^{(l)})$\\
    \vspace{2mm}
    \hspace*{4.3mm}\textcolor{teal}{\# Decompose into magnitude and phase (Cartesian to Polar)}\\
    \hspace*{4.3mm}$\tilde R^{(l)} \leftarrow \sqrt{\Re(\tilde Z^{(l)})^2 + \Im(\tilde Z^{(l)})^2}$\\
    \hspace*{4.3mm}$\tilde \Phi^{(l)} \leftarrow \text{atan2}\left(\Im(\tilde Z^{(l)}), \Re(\tilde Z^{(l)})\right)$\\
    \vspace{2mm}
    \hspace*{4.3mm}\textcolor{teal}{\# Implicitly solve for $R^{(l+1)}$ with Newton's method or Cardano's formula}\\
    \hspace*{4.3mm}Solve $R^{(l+1)} = \tilde R^{(l)} + dt(\alpha - \beta (R^{(l+1)})^3)$\\
    \vspace{2mm}
    \hspace*{4.3mm}\textcolor{teal}{\# Explicitly update the phase using $R^{(l+1)}$}\\
    \hspace*{4.3mm}$\Phi^{(l+1)} \leftarrow \tilde \Phi^l{(l)} + dt(\omega + \gamma (R^{(l+1)})^2)$\\
    \vspace{2mm}
    \hspace*{4.3mm}\textcolor{teal}{\# Recompose into a complex value (Polar to Cartesian)}\\
    \hspace*{4.3mm}$Z^{(l+1)} \leftarrow R^{(l+1)}\cos(\Phi^{(l+1)}) + i \cdot R^{(l+1)}\sin(\Phi^{(l+1)})$\\
\textbf{end for}\\
\vspace{2mm}

\textcolor{teal}{\# Decode the real component of the final layer's features}\\
$Y_{out} \leftarrow \text{DEC}(\Re(Z^{(L)}))$\\
\vspace{2mm}

\textbf{Return} $Y_{out}$
\end{flushleft}
\caption{Pseudocode of \ACRONYM{}.}
\end{algorithm}

The implicit update of $R^{(l+1)}$ can be efficiently computed using either 1D Newton's method or explicit inverse with Cardano's formula, using the real root of the corresponding cubic equation. We discuss the details in Appendix \ref{app:solve_implicit_magnitude}.
When Newton's method is used, it typically converges within 2 - 5 steps, depending on the set error tolerance. Both the encoder {\ttfamily ENC} and the decoder {\ttfamily DEC} are single-layered MLPs. While {\ttfamily ENC} is complex-valued, from our experience the choice between a real- or complex-valued {\ttfamily DEC} does not noticeably affect model performance. Other readout types including modulus, phases, and phase differences do not improve and may sometimes degrade model performance. 
Although this scheme requires repeated polar-cartesian conversion at each layer, empirically this introduces negligible computational overhead and does not impact model performance. This compares favorably with the alternative aforementioned 2D Newton's method, which does not necessitate such conversions but is more computationally intensive. 

\section{Experimental Evaluation}
In this section, we conduct an extensive experimental evaluation of \ACRONYM{} against existing OGNNs and the non-oscillatory baseline. For a fair comparison, we implement all models focusing on the core oscillatory mechanisms, without using additional model-specific artifacts. We extensively tune the hyperparameters of all models presented in this section, with the search spaces defined in Appendix \ref{app:search_space}. We release the best hyperparameters along with the code (available upon publication).

\subsection{Model Setups}

\paragraph{Baselines} 
The baseline models in our study serve as non-oscillatory benchmarks for comparison. These models, which do not incorporate an explicit local oscillator term, also form the basis for the coupling functions in OGNNs. 
The update of these models can be expressed as a forward Euler discretization of an ODE:
\begin{equation}
    X^{(l+1)} = X^{(l)} + dt~F^{(l)}_{\theta}(X^{(l)})
\end{equation}
where $X^{(l)}$ is the matrix of node features at time $t$, $dt$ is the step size, $F^{(l)}_{\theta}(X^{(l)})$ represents the nonlinear coupling with the neighboring nodes based on the graph structure.
To ensure a fair comparison, we explicitly include a skip connection ($X^{(l)}+\dots$) in our baseline implementations, inherited from the Euler discretization. This makes the baselines more architecturally consistent with OGNNs' ODE structure. We highlight that in our experience, adding the skip connection consistently improves model performance.  We consider three coupling functions, following a similar setup of prior works:
\begin{itemize}[topsep=0pt, leftmargin=*,itemsep=0pt]
\item \textbf{GCN}: The Graph Convolutional Network \cite{kipfSemiSupervisedClassificationGraph2016} coupling is defined as $F_{\theta}^{(l)}(X^{(l)}) = \sigma (D^{-\frac{1}{2}}A D^{-\frac{1}{2}}X^{(l)}W^{(l)}
)$, where $\sigma$ is LeakyReLU with tunable negative gradient, $A$ is the adjacency map of the graph augmented by self loops and $D$ is the diagonal matrix such that $D_{ii}=\sum_j A_{ij}$. $W^{(l)}$ is a learnable weight matrix. 
\item \textbf{GAT}: The Graph Attention Network \cite{velickovicGraphAttentionNetworks2017} coupling is defined as $F_{\theta}^{(l)}(X^{(l)}) = \sigma(A^{(l)}(X^{(l)})X^{(l)}W^{(l)})$, where the adjacency matrix is computed with an attention mechanism. The entry for edge $(i,j)$ is given by:
$$A(X^{(l)})_{ij}:=\frac{\text{exp}\Big(\sigma(a^{(l)\top}[W^{(l)}X_i^{(l)}|| W^{(l)}X_j^{(l)}]\Big)}{\sum_{k\in \mathcal{N}_i}\text{exp}\Big(\sigma(a^{(l)\top}[W^{(l)}X_i^{(l)}|| W^{(l)}X_k^{(l)}]\Big)},$$
where $a^{(l)}$ is a learned $2m$-dimensional vector and $X_{i}\in \mathbb{R}^{m}$ is the $i$-th row of the matrix $X$, i.e. the features of the node $i$, and finally $\mathcal{N}_i$ denotes the neighborhood of the node $i$ in the graph~$\mathcal{G}$. 
\item \textbf{Tran}: we also consider a feature-propagation-like setup similar to GRAND \cite{chamberlainGRANDGraphNeural2021}, where the feature update for each node is driven by a weighted sum of differences from its neighbors. Here the coupling function is defined as 
\begin{align}
    F_{\theta}(X^{(l)}) &= \sigma\Big( \kappa\sum_{j\in\mathcal{N}_i}a_\theta^{(l)}(X_i,X_j)(X_j-X_i) \Big), \\
    a_\theta^{(l)}(X_i,X_j) &= \text{softmax}\Big(\frac{(W_K^{(l)}X_i)^\top(W_Q^{(l)}X_j)}{d_k}\Big),
\end{align}
where $\kappa$ is the coupling strength coefficient, $W_Q^{(l)}$ and $W_K^{(l)}$ are learnable query and key matrices, $d_k$ is the key dimension.
\end{itemize}

\paragraph{Second Order Harmonic Oscillator (GraphCON \cite{ruschGraphCoupledOscillatorNetworks2022})} 
In this case, the evolution of the features is governed by a second order ODE and solved with symplectic Euler:
\begin{align*}
Y^{(l+1)} &= Y^{(l)} + dt ~ \sigma(F^{(l)}_{\theta}(X^{(l)}))-\gamma X^{(l)} - \alpha Y^{(l)}, \\
X^{(l+1)} &= X^{(l)} + dt ~ Y^{(l+1)},
\end{align*}
where the coupling $F_{\theta}^{(l)}$ represents the driving force, $\alpha$ models the angular frequency and $\gamma$ the damping force. The variants GraphCON-GCN, GraphCON-GAT, and GraphCON-Tran are instantiated using the respective coupling functions $F_\theta^{(l)}$ described previously.

\paragraph{Kuramoto}
The Kuramoto model treats the features of each node as the phase constrained on the surface of a hypersphere. The following Kuramoto architecture is proposed in \cite{nguyenCoupledOscillatorsGraph2023}
\begin{equation} \label{eq:kuramotognn}
    X_i^{(l+1)}= X_i^{(l)} + dt\Big(\omega+ \kappa\sum_{k\in \mathcal{N}_i} a_\theta^{(l)}(X_i,X_k)\mathrm{sin}(X_{k}^{(l)}-X_{i}^{(l)}) \Big),
\end{equation}
where $\omega$ is the natural frequency, $\mathrm{sin}(X_{k}^{(l)}-X_{i}^{(l)})$ the nonlinear coupling term, and $a_\theta^{(l)}$ represents a learnable attention mechanism.
However, directly solving Equation \eqref{eq:kuramotognn} with explicit Euler is implementation-wise challenging, as it is difficult to constrain the phase features within the $[0, 2\pi]$ interval, leading to instability during backpropagation and significant underperformance. While prior work \cite{nguyenCoupledOscillatorsGraph2023} used adaptive step-size solvers, these can be computationally expensive and prone to step-size underflow. To address this, as well as improving the adaptability of the model to different coupling functions, we propose an alternative numerical scheme by representing the phases as complex numbers on the unit circle $Z=e^{\mathbf{i}X}$, and rewrite the layer update as the following
\begin{align*}
    \tilde Z^{(l)} &= Z^{(l)} + dt~F^{(l)}_\theta(Z^{(l)}), \\
    \Phi^{(l+1)} &= \text{atan2}\left(\Im(\tilde Z^{(l)}), \Re(\tilde Z^{(l)})\right) + dt ~ \omega, \\
    Z^{(l+1)} &= \cos(\Phi^{(l+1)}) + \sin(\Phi^{(l+1)})i,
\end{align*}
where atan2 is the two-argument arctangent function. This is analogous to the IMEX scheme used for \ACRONYM{}, but without implicitly solving the magnitude term. Compared to adaptive step-size solvers like Dormand-Prince, this approach yields a runtime reduction of up to 100$\times$, while achieving comparable or better performance. We also highlight that the proposed approach directly incorporates a generic learnable coupling function $F_\theta^{(l)}$, as opposed~to~\eqref{eq:kuramotognn}.

\begin{table}[t]
\centering
\resizebox{\columnwidth}{!}{
\input{Tables/homophilic}
}
\caption{Transductive node classification accuracy (\%) on homophilic datasets. The means and standard deviations are computed over 100 random dataset splits and initialization. \highlight{darkgray}{Dark gray} and \highlight{lightgray}{light gray} indicate the best and second best values, respectively.}
\label{tab:homophilic}
\end{table}

\subsection{Node Classification}
We evaluate all models on transductive node classification tasks on both homophilic and heterophilic datasets, including Cora \cite{mccallumAutomatingConstructionInternet2000}, Citeseer \cite{senCollectiveClassificationNetwork2008}, Pubmed \cite{namataQuerydrivenActiveSurveying2012}, WebKB \cite{peiGeomGCNGeometricGraph2020} (Texas, Wisconsin, Cornell), WikipediaNetwork \cite{rozemberczkiMultiscaleAttributedNode2019} (Chameleon, Squirrel), heterophilous graphs \cite{platonovCriticalLookEvaluation2023} (Amazon-rating). For homophilic graphs, we follow the data split schemes laid out by \citeauthor{shchurPitfallsGraphNeural2018} \cite{shchurPitfallsGraphNeural2018}. As shown in Table \ref{tab:homophilic} and \ref{tab:heterophilic}, \ACRONYM{} variants consistently achieve competitive or state-of-the-art performance across both dataset types. Statistical significance tests are provided in Appendix \ref{app:t_tests}. 

\begin{table}[t]
\centering
\resizebox{\columnwidth}{!}{
\input{Tables/heterophilic}
}
$\,$\\
\resizebox{\columnwidth}{!}{
\input{Tables/heterophilic2}
}
\caption{Transductive node classification accuracy (\%) on heterophilic datasets. The means and standard deviations are computed over 10 pre-defined splits, with 10 random initializations per split.}
\label{tab:heterophilic}
\end{table}

\subsection{Graph-level tasks}
To further evaluate performance of \ACRONYM{}s on graph-level tasks, We benchmark graph classification performance on the MoleculeNet datasets \cite{wuMoleculeNetBenchmarkMolecular2017} (MUTAG, ENZYMES, PROTEINS), and graph regression on datasets from the Open Graph Benchmark \cite{ivanovUnderstandingIsomorphismBias2019, huOpenGraphBenchmark2020} (ESOL, FreeSolv, Lipophilicity). For these experiments, we adopt a standard 80:10:10 train-validation-test split with a batch size of 64. As presented in Table \ref{tab:graph_classification} and \ref{tab:graph_regression}, the results demonstrate the strong performance of \ACRONYM{}s on these tasks, which consistently outperform other OGNNs by up to 80\% (see ENZYMES-Tran cells).

\begin{table}[t]
\centering
\resizebox{\columnwidth}{!}{
\input{Tables/graph_classification}
}
\caption{Graph classification accuracy (\%) on MoleculeNet \cite{wuMoleculeNetBenchmarkMolecular2017} datasets. The means and standard deviations are averaged over 50 random splits and weight initializations.}
\label{tab:graph_classification}
\end{table}

\begin{table}[t]
\centering
\resizebox{\columnwidth}{!}{
\input{Tables/graph_regression_short}
}
\caption{Graph regression (L2 loss) on OGBG-MOL datasets \cite{ivanovUnderstandingIsomorphismBias2019, huOpenGraphBenchmark2020}. The means and standard deviations are averaged over 50 random splits and weight initializations.}
\label{tab:graph_regression}
\end{table}

\section{Discussions}\label{sec:discussion}
\paragraph{Performance of deep GNNs}
To assess the ability of different OGNNs to construct deep networks and mitigate oversmoothing, we evaluate their performance on Cora with GCN coupling across OGNNs of different depths. We note that the best hyperparameter configurations used in Table \ref{tab:homophilic}-\ref{tab:graph_regression} may not translate to decent performance for deeper architectures. To ensure a fair comparison, we therefore re-tune the models for seven fixed depths ranging from 2 to 128 layers. 

\begin{figure}
    \centering
    \includegraphics[width=0.9\linewidth]{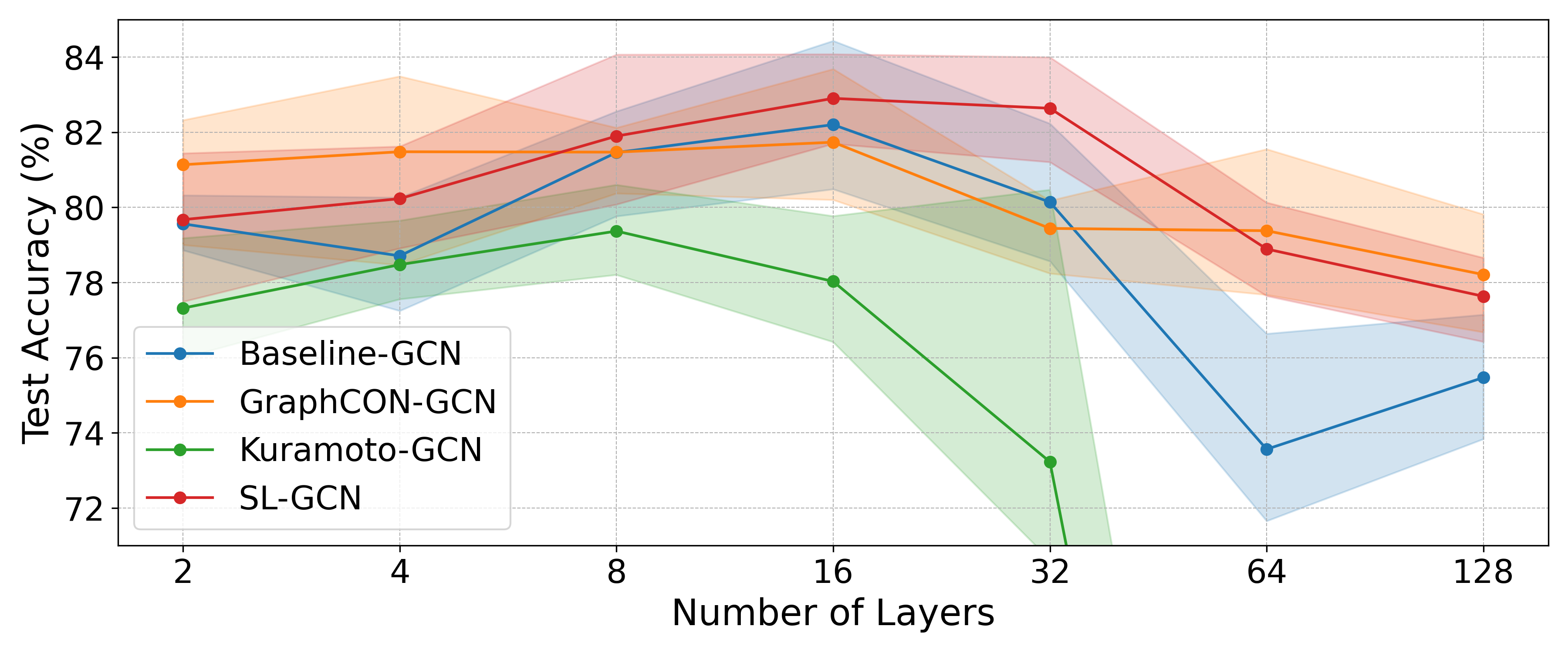}
    \caption{Node classification accuracy (\%) on Cora for GNNs of varying depths. 
    Shaded regions indicate the 10th to 90th percentile range. The mean values are computed over 10 random splits and weight initializations.}
    \label{fig:depth_accuracy}
\end{figure}

Figure \ref{fig:depth_accuracy} shows that SL-GCN maintains high performance even at significant depths, with its accuracy peaking at 32 layers. In contrast, the performances of both GraphCON and Kuramoto-GCN peak at 8 layers, with the latter rapidly collapsing to $29.98 \pm 2.02~\%$ accuracy at 128 layers, indicating its weakness in deep settings. 
These findings align with our hyperparameter search for Table \ref{tab:homophilic}-\ref{tab:graph_regression}, where the optimal configurations for \ACRONYM{} frequently involved deeper layers compared to GraphCON, Kuramoto, and the non-oscillatory baseline. This ability to effectively leverage network depth provides an intuition for the superior performance of \ACRONYM{}s over many evaluated tasks. Furthermore, it is also noteworthy that a simple Baseline-GCN with a skip connection component (added for consistency with the neural ODE framework but often omitted in the literature) can often achieve highly competitive performance against some of the more sophisticated OGNNs, even in very deep settings. Further discussion across model variants and datasets is presented in Appendix \ref{app:depth}.

\begin{figure}
    \centering
\includegraphics[width=0.9\linewidth]{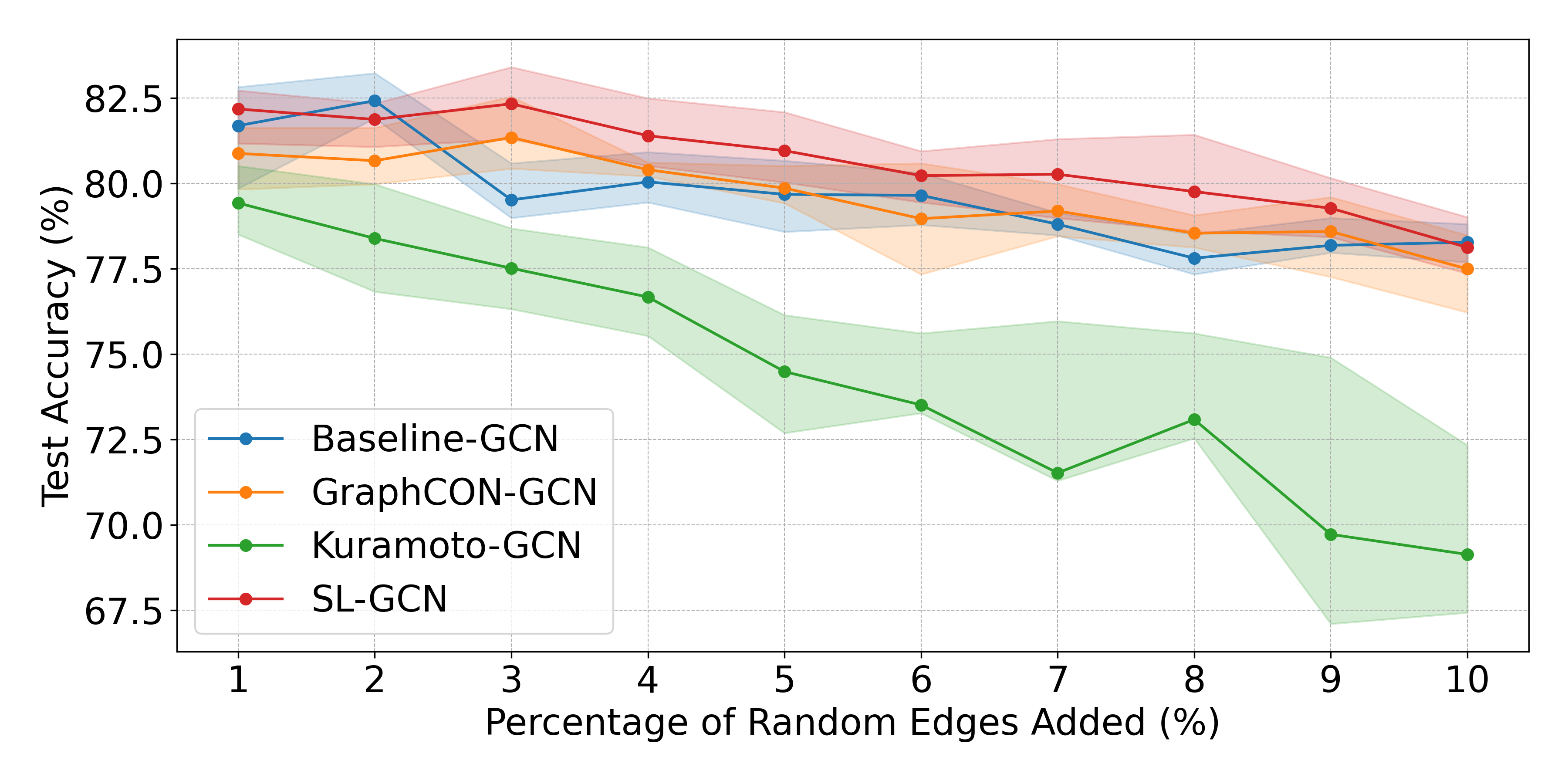}
    \caption{Robustness of GNNs on Cora dataset against random edge perturbations. Shaded regions indicate the 25th to 75th percentile range, computed over 10 random splits and weight initializations.}
    \label{fig:robustness}
\end{figure}

\paragraph{Robustness to graph perturbations}
Beyond depth scalability, we also evaluate model robustness against graph perturbations by randomly adding fake edges to the graph structure, following the protocol from \citeauthor{fengGraphRandomNeural2020} \cite{fengGraphRandomNeural2020}. As shown in Figure \ref{fig:robustness}, \ACRONYM{} exhibits superior robustness compared to Kuramoto and achieves a level of robustness comparable to that of GraphCON. 

\paragraph{Criticality of the Stuart Landau oscillator}
A key property of the Stuart Landau Oscillator is its ability to operate near a critical state, which has been extensively used by the the neuroscience community for modeling the brain \cite{deco}. At this critical point, the system's dynamics converge at a slower, algebraic rate rather than an exponential one. This state is reached when the following condition is met: for each node $j$,
\begin{align}\label{critcond}
\alpha+\kappa\sum_{k=1}^N A_{jk} (cos(\phi_j-\phi_k)\frac{r_k}{r_j}-1)=0.
\end{align}
Directly verifying this condition is challenging, as the coupling strength $\kappa$ is implicitly set by the magnitude of the learned weights for SL-GCN and -GAT, and $A_{jk}$ is dependent on the attention mechanism for SL-GAT and -Tran, both of which may not be fixed across layers. However, we find compelling empirical evidence that the model can leverage this property and reach a near-critical state. For instance, in our experiments with SL-GCN on Cora, Citeseer, and Pubmed, the hyperparameter tuner consistently selects a Hopf-parameter $\alpha$ close to 0 (e.g. 0.04), while we concurrently observe the average absolute value of $cos(\phi_j-\phi_k)\frac{r_k}{r_j}$ remains around 1.0 (e.g. 1.06). In this special case, the oscillators appear to be almost synchronized, but with $\alpha \sim 0$, the Stuart-Landau oscillator is near its critical state. 
This suggests the hyperparameter tuner has leveraged the slower algebraic convergence to improve the performance of the model, a characteristic that no other OGNN frameworks have. 
This type of criticality-seeking behaviour appears to align with the idea that the brain operates near criticality \cite{RevModPhys.90.031001}. The precise implications in deep learning will be the topic of a future study.

\paragraph{Computational Cost} A practical consideration for \ACRONYM{} is its computational cost. The bespoke IMEX time-stepping scheme, necessary to handle the stiff cubic term in the Stuart-Landau dynamics, incurs an overhead compared to the fully explicit methods used for Kuramoto and GraphCON models. Consequently, \ACRONYM{}'s training times are typically about double those of other OGNN counterparts.
However, our IMEX implementation is highly optimized: by reducing the implicit part to a one-dimensional problem, it achieves a runtime improvement by over two orders of magnitude compared to standard adaptive solvers like the Dormand-Prince method. This approach not only makes the model computationally feasible but also circumvents issues like time-step underflow while maintaining excellent numerical stability.

\section{Conclusion}
In this work, we introduced the Stuart-Landau Graph Neural Network (\ACRONYM{}), a novel oscillatory architecture grounded in the rich dynamics of Stuart-Landau systems. By modeling the coupled evolution of both feature amplitude and phase, \ACRONYM{} generalizes prior phase-only models and gains significant expressive power. Aided by a bespoke implicit-explicit numerical scheme for stable and efficient training, our extensive experiments on node, graph, and regression tasks demonstrate that this Stuart-Landau-based approach translates into state-of-the-art performance, outperforming existing oscillatory GNNs. 

\section{Acknowledgement}
KZ was supported by the EPSRC Centre for Doctoral Training in Mathematical Modelling, Analysis and Computation (MAC-MIGS) funded by the UK Engineering and Physical Sciences Research Council (grant EP/S023291/1), Heriot-Watt University and the University of Edinburgh. DNR was funded by National Science Centre, Poland, grant number 2023/50/A/ST1/00447 and partially supported by the Modeling Nature (MNat) Research Unit, project QUAL21-011. PD and FT are members of the Gruppo Nazionale Calcolo Scientifico - Istituto Nazionale di Alta Matematica (GNCS-INdAM). 
FT is partially funded by the PRIN-MUR project MOLE (code 2022ZK5ME7) and the PRIN-PNRR project FIN4GEO (code P2022BNB97). PD was supported by the INdAM-GNCS project “NLA4ML—Numerical Linear Algebra Techniques for Machine Learning”. This work was supported by the Edinburgh International Data Facility (EIDF) and the Data-Driven Innovation Programme at the University of Edinburgh, as well as AI Research Resource Programme (AIRR) under UK Research and Innovation (UKRI).

\bibliographystyle{ACM-Reference-Format}
\balance
\bibliography{DRAFT/bibliography}

\newpage
\appendix

\section{Hyperparameter Tuning Setup} \label{app:search_space}
We conduct extensive hyperparameter search across all model-dataset pairs presented in Table \ref{tab:homophilic}-\ref{tab:graph_regression} with Ray platform \cite{moritzRayDistributedFramework2017} and Optuna \cite{akibaOptunaNextgenerationHyperparameter2019} search algorithm over 1000 trials for node-level tasks and 500 for graph-level ones. The search spaces for all GCN, GAT, and Tran variants include: latent dimension size \texttt{[16,128]}, number of layers \texttt{[1,80]}, learning rate \texttt{[10$^{-5}$, 10$^{-2}$]}, weight decay \texttt{[10$^{-6}$, 1.0]}, LeakyReLU gradient \texttt{[0.0, 1.0]}, dropout and input dropout rates \texttt{[0.0, 0.5]}. The time step size is fixed at $dt=$ \texttt{1.0}. For GAT and Tran, we fix the number of attention heads to 2 or 4. For Tran, we additionally tune the attention dimension between \texttt{[4, 80]} and the coupling strength $\kappa$ in \texttt{[0.1,1.5]}. 
For the OGNNs, we tuned their specific parameters:
\begin{itemize}[topsep=0pt, leftmargin=*,itemsep=0pt]
    \item For \ACRONYM{}, we search $\alpha$ in \texttt{[-0.5,0.5]}, $\beta$ in  \texttt{[0.0,2.0]}, $\omega$ in \texttt{[0.5,2.0]} and $\gamma$ in \texttt{[-1.0,2.0]}. 
    \item For Kuramoto, we tune the natural frequency $\omega$ in \texttt{[0.0, 2.0]}, which is consistent with that of \ACRONYM{}.
    \item For GraphCON, we tune both $\alpha$ and $\gamma$ in \texttt{[0.0, 1.5]}, consistent with the search space used in \cite{ruschGraphCoupledOscillatorNetworks2022}.
\end{itemize}
Hyperparameters are optimized by maximizing classification accuracy or minimizing regression loss on the validation set. We report the final evaluation results on the test set with best-performing configuration on the validation set.

\section{Statistical Significance Tests}\label{app:t_tests}
To validate the results presented in Table \ref{tab:homophilic}-\ref{tab:graph_regression}, we conduct one-tailed t-tests to assess the statistical significance of the performance improvements achieved by our \ACRONYM{} models. We use a 95\% confidence level for all tests.  
For each dataset and model variant (GCN, GAT, Tran), we compare the best-performing \ACRONYM{} against the second best-performing model. In the few cases where \ACRONYM{} is the second-best performer, we compare it against the third-best one. The t-score is calculated as follows:
$$\text{t-score} = \frac{\mu_\text{SL}-\mu_\text{other}}{\sqrt{\frac{\sigma^2_\text{SL}}{n}+\frac{\sigma^2_\text{other}}{n}}},$$
where $\mu$ and $\sigma^2$ represent the mean and variance of the model performance over $n$ number of repeated trials. 
A t-score greater than 1.66 indicates that \ACRONYM{}'s performance is significantly superior with 95\% confidence. As presented in Table \ref{tab:t_test_homophilic}, the results show that \ACRONYM{} achieve statistically significant gains in many model variant-dataset pairs, providing strong evidence for the effectiveness of the proposed architecture.

\begin{table}[h]
\centering
\resizebox{\columnwidth}{!}{
\input{Tables/t-test_homophilic}
}
{\setlength{\tabcolsep}{0.7mm}
\resizebox{\columnwidth}{!}{
\input{Tables/t-test_homophilic2}
}}
\resizebox{0.6\columnwidth}{!}{
\input{Tables/t-test_homophilic3}
}
\caption{T-test scores of the results in Table \ref{tab:homophilic}-\ref{tab:graph_regression}. For regression tasks (ESOL, FreeSolv, Lipophilicity), we present the negative t-score.}
\label{tab:t_test_homophilic}
\end{table}

\section{Additional Results on Model Depths}\label{app:depth}
\begin{figure}[h]
    \centering
    \includegraphics[width=0.95\columnwidth]{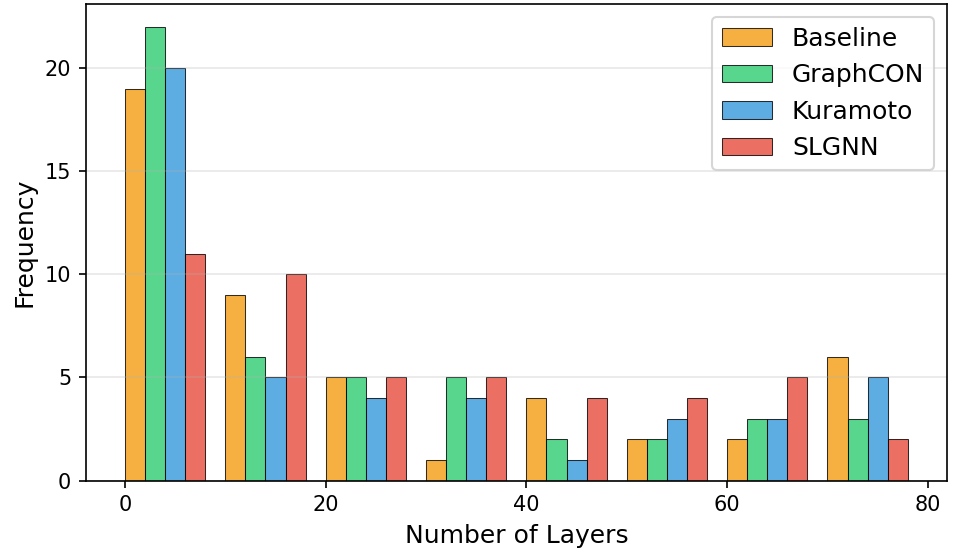}
    \caption{Distribution of optimal layer depths selected by the hyperparameter tuner for different model types.}
    \label{fig:nlayer_distribution}
\end{figure}

This appendix extends the discussion from Section \ref{sec:discussion} on the relationship between network depth and model performance. In Figure \ref{fig:nlayer_distribution}, we present the distribution of the optimal model depths selected by hyperparameter tuning algorithm. 

We note that \ACRONYM{}'s optimal configurations are less concentrated in the shallow 1- to 10-layer range compared to GraphCON, Kuramoto, and the Baseline. This reinforces our claim that \ACRONYM{}s typically leverage deeper network structures to achieve better performance. On the other hand, there are a significant number of Kuramoto's configurations in the very deep 70- to 80-layer range. At first glance, this may seem to contradict the conclusions drawn from Figure \ref{fig:depth_accuracy}. However, we observe that models with optimal depths in this range are predominantly those using Tran coupling, a configuration that makes models more robust at greater depths comparing to the GCN or GAT coupling. It is also known that certain datasets, such as the heterophilic ones, are less susceptible to performance degradation in deep architectures \cite{guoTamingOversmoothingRepresentation2023}. Most importantly, increased depth does not always translate to better performance, particularly for the Baseline and other OGNN models.

To quantify this relationship, we calculated the correlation coefficient between network depth used in the best hyperparameter configuration and model accuracy across all node and graph classification tasks in Table \ref{tab:nlayer_acc_corr_coeff}. The results show that \ACRONYM{} has a substantially stronger positive correlation between depth and performance than other model classes. This indicates that SLGNN is more effective at leveraging additional layers for performance gains. Therefore, while it may not always require maximum depth, its ability to effectively use deeper architectures is superior, again reinforcing the conclusions from our main analysis.

\begin{table}[h]
    \centering
    \input{Tables/nlayer_corr_coeff}
    \caption{Correlation coefficients between the optimal layer depths and the model accuracies on the node and graph classification tasks from Table \ref{tab:homophilic}-\ref{tab:graph_classification}. For each dataset, the accuracy is normalized by the worst performing accuracy from any model to set a performance baseline of 1.}
    \label{tab:nlayer_acc_corr_coeff}
\end{table}

\section{Implicit Solving Magnitude Equation} \label{app:solve_implicit_magnitude}
This appendix discusses using Newton's method  or explicit inverse to implicitly solve the magnitude component in Algorithm 1, 
\begin{align}
    R^{(l+1)} &= \tilde R^{(l)} + dt(\alpha - \beta (R^{(l+1)})^3).
\end{align}

For Newton, one can define 
\begin{align*}
    F(R^{(l+1)}, \tilde R^{(l)}) &= R^{(l+1)} - \tilde R^{(l)} - dt\dot r(R^{(l+1)}) = 0, \\
    \frac{\partial F(R^{(l+1)},\tilde R^{(l)})}{\partial R^{(l+1)}} &= 1 - dt(\alpha - 3\beta (R^{(l+1)})^2).
\end{align*}
The Newton update can then be expressed as 
\begin{gather*}
    R^{(l+1)} \leftarrow R^{(l+1)} - \left( \frac{\partial F(R^{(l+1)},\tilde R^{(l)})}{\partial R^{(l+1)}} \right)^{-1} F(R^{(l+1)}, \tilde R^{(l)}).
\end{gather*}
The iteration stops when the maximum component of subsequent update is less than a pre-set tolerance. We typically set this tolerance to $10^{-5}$, although larger values can also be used to marginally reduce runtime without noticeable drop in model performance. 

For explicit inverse, we re-write the above equation in its depressed form 
\begin{gather*}
    \left(R^{(l+1)}\right)^3 + \left(\frac{1-dt\alpha}{dt\beta}\right)R^{(l+1)} + \left(\frac{-\tilde R^{(l)}}{dt\beta}\right) = 0,
\end{gather*}
where we let $p=(1-dt\alpha)/(dt\beta)$, $q=-\tilde R^{(l)}/(dt\beta)$, and define
$$u_\pm = -\frac q2\pm\sqrt{\left(\frac q2\right)^2+\left(\frac p3\right)^3}, \qquad \epsilon_\pm=\dfrac{-1\pm i\sqrt3}2,$$
where $\epsilon_\pm$ are the roots of unity. The three roots of the cubic equation are then given by Cardano's formula: $\sqrt[3]{u_+}+\sqrt[3]{u_-}$ (always real), $\epsilon_+\sqrt[3]{u_+}+\epsilon_-\sqrt[3]{u_-}$ and $\epsilon_+\sqrt[3]{u_-}+\epsilon_-\sqrt[3]{u_+}$. The latter two roots are real if the condition $(q/2)^2+(p/3)^3>0$ is satisfied. In most cases, this equation typically only has one real root, unless the feature values are exploding due to a poorly chosen set of hyperparameters.

\section{Computational Complexity}
The computational complexity of the proposed \ACRONYM{} is consisted of the coupling part and the IMEX part. The coupling complexity depends on the specific coupling function used. For instance, a GCN coupling scales as $\mathcal{O}(|E|d+|V|d^2)$, where $|V|$ and $|E|$ denote the number of nodes and edges, respectively, and $d$ represents the feature dimension.

Regarding the IMEX scheme, the implicit step using Newton's method has a per-feature, per-iteration complexity of $\mathcal{O}(1)$. Consequently, the complexity of the IMEX scheme over the entire graph is $\mathcal{O}(k|V|d)$, where $k$ represents the number of Newton iterations. Our empirical observations suggest that $k$ depends primarily on the feature distribution rather than the graph size. 

Empirically, SLGNN is observed to be up to $2\times$ slower per epoch than Kuramoto and up to $3\times$ slower than GraphCON for the best hyperparameter configurations as shown in Table \ref{tab:runtime}. However, it remains substantially faster than the explicit high-order ODE solvers used in prior works, such as the Dormand Prince method. In terms of memory efficiency, the VRAM usage of SLGNN is comparable to Kuramoto and within $1.5\times$ of GraphCON as shown in Table \ref{tab:vram}. It is important to note that SLGNN often leverages deeper architectures to achieve optimal performance as shown in Figure \ref{fig:nlayer_distribution}, which naturally contributes to increased computation time. This additional computational cost is counterbalanced by the model's superior predictive performance.

\begin{table}[h]
\centering
\caption{Runtime per epoch (ms) comparison. Values represent mean $\pm$ standard deviation.}
\label{tab:runtime}
\resizebox{\columnwidth}{!}{
\begin{tabular}{c c c c}
\toprule
Model & Cora & Citeseer & Pubmed \\
\midrule
Baseline-GCN & $9.62 \pm 0.74$ & $11.10 \pm 0.38$ & $18.39 \pm 0.55$ \\
GraphCON-GCN & $11.94 \pm 0.51$ & $13.68 \pm 1.13$ & $20.42 \pm 0.56$ \\
Kuramoto-GCN & $16.22 \pm 0.48$ & $19.00 \pm 0.31$ & $24.73 \pm 0.41$ \\
SL-GCN & $32.82 \pm 2.32$ & $32.40 \pm 1.10$ & $41.22 \pm 1.79$ \\
\bottomrule
\end{tabular}
}
\resizebox{\columnwidth}{!}{
\begin{tabular}{c c c c}
\toprule
Model & Texas & Wisconsin & Cornell \\
\midrule
Baseline-GCN & $10.40 \pm 0.72$ & $11.59 \pm 0.53$ & $11.16 \pm 0.49$ \\
GraphCON-GCN & $12.92 \pm 0.57$ & $13.73 \pm 0.24$ & $13.43 \pm 0.57$ \\
Kuramoto-GCN & $18.67 \pm 0.81$ & $18.76 \pm 0.40$ & $18.95 \pm 0.70$ \\
SL-GCN & $37.59 \pm 1.89$ & $36.98 \pm 1.20$ & $39.85 \pm 1.85$ \\
\bottomrule
\end{tabular}
}
\end{table}

\begin{table}[h]
\centering
\caption{Peak VRAM Usage (MB) comparison. Values represent mean $\pm$ standard deviation.}
\label{tab:vram}
\resizebox{\columnwidth}{!}{
\begin{tabular}{c c c c}
\toprule
Model & Cora & Citeseer & Pubmed \\
\midrule
Baseline-GCN & $66.42 \pm 0.00$ & $98.50 \pm 0.00$ & $262.79 \pm 0.00$ \\
GraphCON-GCN & $68.26 \pm 0.00$ & $97.53 \pm 0.00$ & $278.06 \pm 0.00$ \\
Kuramoto-GCN & $129.34 \pm 0.00$ & $259.35 \pm 0.00$ & $390.16 \pm 0.00$ \\
SL-GCN & $123.06 \pm 2.74$ & $179.64 \pm 5.61$ & $574.39 \pm 10.20$ \\
\bottomrule
\end{tabular}
}
\resizebox{\columnwidth}{!}{
\begin{tabular}{c c c c}
\toprule
Model & Texas & Wisconsin & Cornell \\
\midrule
Baseline-GCN & $22.92 \pm 0.00$ & $26.75 \pm 0.00$ & $27.14 \pm 0.00$ \\
GraphCON-GCN & $22.42 \pm 0.00$ & $26.19 \pm 0.00$ & $27.31 \pm 0.00$ \\
Kuramoto-GCN & $27.41 \pm 0.00$ & $33.21 \pm 0.00$ & $32.30 \pm 0.00$ \\
SL-GCN & $31.32 \pm 2.14$ & $36.12 \pm 2.55$ & $34.75 \pm 1.41$ \\
\bottomrule
\end{tabular}
}
\end{table}

\end{document}

%% file: Tables/oscillator_property_comparison_v3.tex
\newcommand{\myfigurewidth}{0.11\textwidth}

\resizebox{2\columnwidth+0.8cm}{!}{
\begin{tabular}{l c c c c c c c}
\toprule
\multirow{7}{*}{\bf{Oscillator Type}} & \multicolumn{3}{c}{\bf{Magnitude Behaviour}} & \multicolumn{2}{c}{\bf{Magnitude Convergence}} & \multicolumn{2}{c}{\bf{Phase Behaviour}} \\
\cmidrule(lr){2-4} \cmidrule(lr){5-6} \cmidrule(lr){7-8} 
& Amplitude Death & Orbit & Amplitude Growth & Algebraic & Exponential & Incoherence & Synchronized \\
& 
\adjustbox{valign=c}{\includegraphics[width=\myfigurewidth, trim={15pt 15pt 15pt 15pt}, clip]{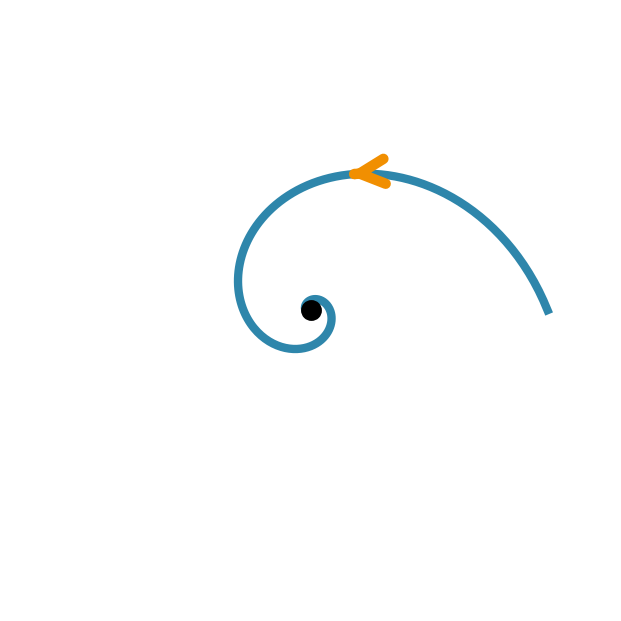}} & 
\adjustbox{valign=c}{\includegraphics[width=\myfigurewidth, trim={15pt 15pt 15pt 15pt}, clip]{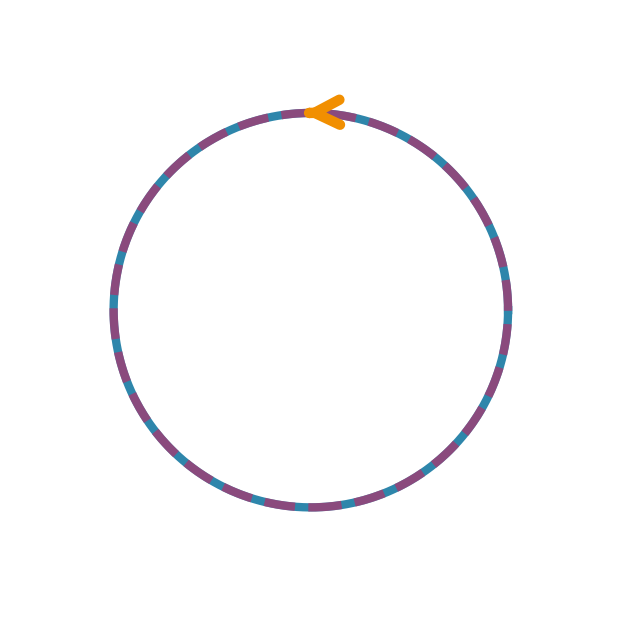}} & 
\adjustbox{valign=c}{\includegraphics[width=\myfigurewidth, trim={15pt 15pt 15pt 15pt}, clip]{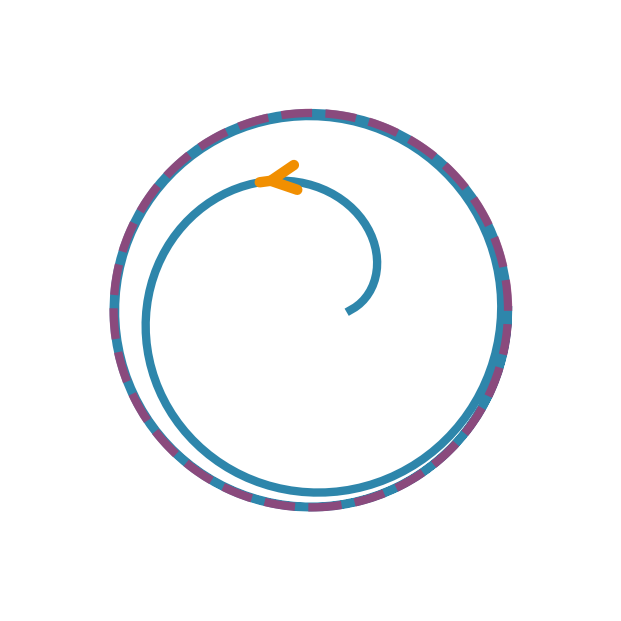}} & 
\adjustbox{valign=c}{\includegraphics[width=\myfigurewidth, trim={15pt 15pt 15pt 15pt}, clip]{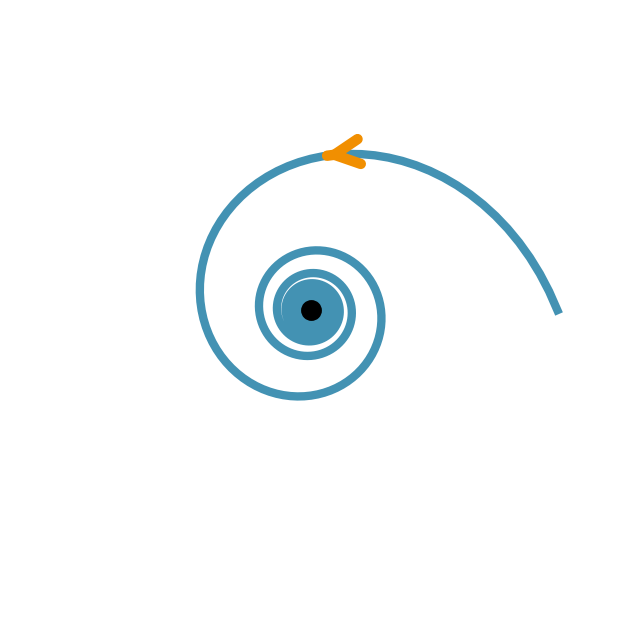}} & 
\adjustbox{valign=c}{\includegraphics[width=\myfigurewidth, trim={15pt 15pt 15pt 15pt}, clip]{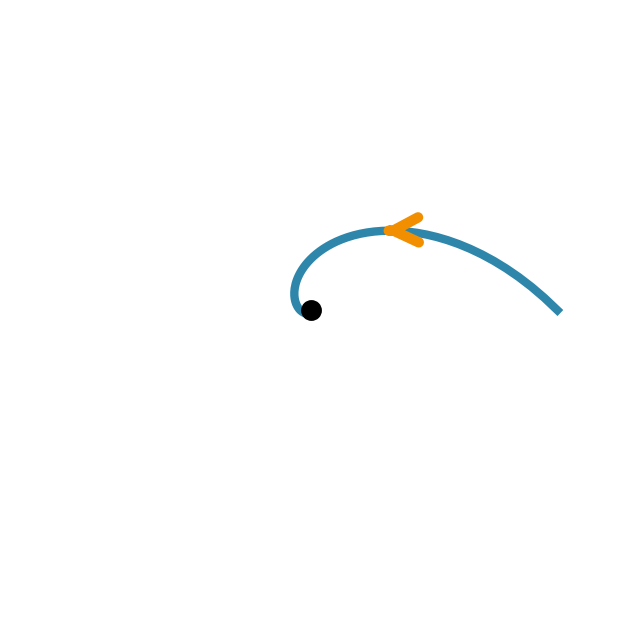}} & 
\adjustbox{valign=c}{\includegraphics[width=\myfigurewidth, trim={0pt 16pt 30pt 16pt}, clip]{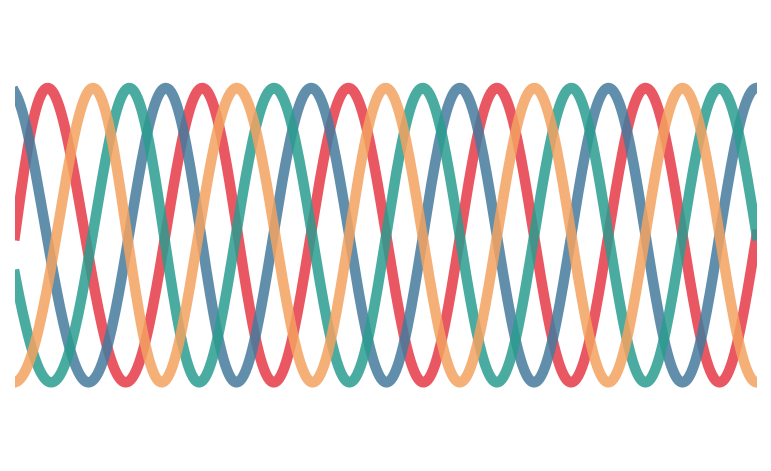}} & 
\adjustbox{valign=c}{\includegraphics[width=\myfigurewidth, trim={0pt 16pt 30pt 16pt}, clip]{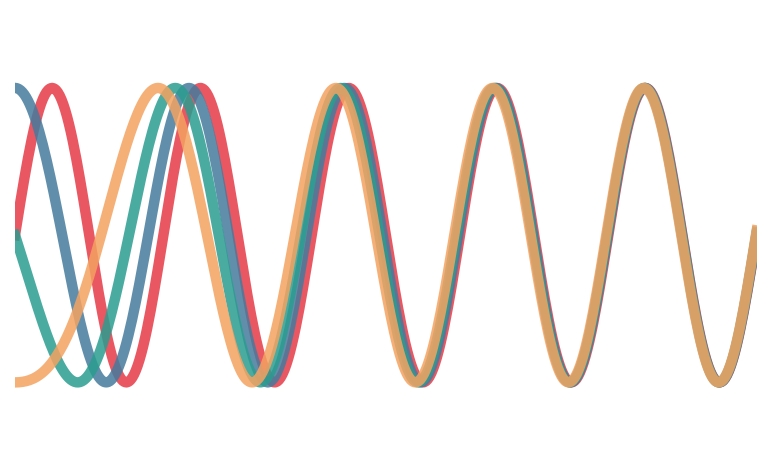}} \\
\midrule
Non-oscillatory & \cmark & \xmark & \xmark & \xmark & \xmark & \xmark & \xmark\\ 
Harmonic & \cmark & \cmark & \xmark & \xmark & \cmark & \cmark & \cmark\\ 
Kuramoto & \xmark & \cmark & \xmark & \xmark & \xmark & \cmark & \cmark \\ 
Stuart Landau & \cmark & \cmark & \cmark & \cmark & \cmark & \cmark & \cmark \\
\bottomrule
\end{tabular}
}

%% file: Tables/homophilic.tex
\begin{tabular}{c c c c}
\toprule
Model & Cora & Citeseer & Pubmed \\
\midrule 
Baseline-GCN & \highlight{lightgray}{82.35 ± 1.61} & \highlight{lightgray}{73.16 ± 1.94} & \highlight{lightgray}{78.24 ± 1.54} \\
GraphCON-GCN & 81.50 ± 1.57 & 72.83 ± 1.80 & 77.95 ± 1.30 \\
Kuramoto-GCN & 79.46 ± 2.05 & 70.83 ± 2.13 & 76.38 ± 3.72\\
SL-GCN & \highlight{darkgray}{82.92 ± 1.39} & \highlight{darkgray}{73.29 ± 2.04} & \highlight{darkgray}{79.21 ± 1.30}\\ 
\midrule
Baseline-GAT & 80.41 ± 1.76 &\highlight{darkgray}{72.95 ± 1.65}&77.31 ± 1.67\\
Graphcon-GAT & \highlight{lightgray}{81.64 ± 1.69} & 72.60 ± 1.79& 77.33 ± 1.62 \\
Kuramoto-GAT & 80.63 ± 1.53 & 71.41 ± 2.08 & \highlight{lightgray}{77.38 ± 2.20} \\
SL-GAT & \highlight{darkgray}{81.97 ± 1.67} & \highlight{lightgray}{72.93 ± 1.79} &\highlight{darkgray}{77.55 ± 1.64} \\
\midrule
Baseline-Tran &\highlight{darkgray}{83.27 ± 1.29} & \highlight{lightgray}{73.56 ± 1.78} & \highlight{lightgray}{78.60 ± 1.67} \\
Graphcon-Tran & 82.11 ± 1.41 & 72.51 ± 1.94 & 76.64 ± 1.85 \\
Kuramoto-Tran & 82.29 ± 1.22 & 72.97 ± 2.16 &  78.46 ± 1.45 \\
SL-Tran & \highlight{lightgray}{82.84 ± 1.50} & \highlight{darkgray}{74.43 ± 1.71} & \highlight{darkgray}{79.02 ± 1.76}\\
\bottomrule
\end{tabular}

%% file: Tables/heterophilic.tex
\begin{tabular}{c c c c}
\toprule
Model & Texas & Wisconsin & Cornell \\
\midrule
Baseline-GCN &79.43 ± 5.58&84.39 ± 4.36&73.70 ± 4.17\\
GraphCON-GCN & 78.49 ± 5.85 & 83.16 ± 4.43 & 71.97 ± 4.58\\
Kuramoto-GCN & \highlight{lightgray}{80.57 ± 5.55} & \highlight{lightgray}{85.82 ± 3.82} & \highlight{lightgray}{74.11 ± 4.38}\\
SL-GCN & \highlight{darkgray}{81.81 ± 5.31} & \highlight{darkgray}{86.08 ± 4.45} & \highlight{darkgray}{74.46 ± 4.36} \\
\midrule
Baseline-GAT &79.00 ± 6.04&84.12 ± 4.30&72.92 ± 4.10\\
Graphcon-GAT &79.62 ± 6.22 &82.73 ± 4.32&72.81 ± 3.95 \\
Kuramoto-GAT & \highlight{darkgray}{81.73 ± 5.67} & \highlight{lightgray}{85.27 ± 4.26} & \highlight{lightgray}{74.19 ± 4.30}\\
SL-GAT & \highlight{lightgray}{81.32 ± 5.41} & \highlight{darkgray}{85.96 ± 4.26}&\highlight{darkgray}{74.27 ± 3.91}\\
\midrule
Baseline-Tran & 75.76 ± 6.44 &83.08 ± 4.74 & 70.16 ± 5.85 \\
Graphcon-Tran & \highlight{lightgray}{82.89 ± 5.14} & 82.63 ± 4.87 & \highlight{lightgray}{72.73 ± 4.64}\\
Kuramoto-Tran & 78.70 ± 5.79 & \highlight{lightgray}{85.04 ± 4.48} & 72.43 ± 4.89\\
SL-Tran & \highlight{darkgray}{85.16 ± 6.06} & \highlight{darkgray}{85.24 ± 4.44}&  \highlight{darkgray}{76.27 ± 5.04}\\
\bottomrule
\end{tabular}

%% file: Tables/heterophilic2.tex
\begin{tabular}{c c c c}
\toprule
Model & Chameleon & Squirrel & Amazon-rating \\
\midrule
Baseline-GCN & 45.02 ± 2.20 & 34.52 ± 1.45 & 46.65 ± 1.86 \\
GraphCON-GCN & 47.83 ± 2.06 & \highlight{lightgray}{35.05 ± 1.46} & 48.07 ± 0.94 \\
KuramotoGNN & \highlight{lightgray}{49.39 ± 2.26} & 34.68 ± 1.27 & \highlight{lightgray}{48.65 ± 0.80} \\
SL-GCN & \highlight{darkgray}{49.77 ± 2.60} & \highlight{darkgray}{35.67 ± 1.49} & \highlight{darkgray}{51.27 ± 0.60} \\
\midrule
Baseline-GAT & 46.58 ± 2.99 & 35.48 ± 1.59 & 49.83 ± 0.65 \\
Graphcon-GAT & 47.88 ± 2.60 & \highlight{darkgray}{36.35 ± 1.64} & 49.22 ± 1.31 \\
Kuramoto-GAT & \highlight{darkgray}{49.25 ± 2.13} & 34.53 ± 1.40 & \highlight{lightgray}{51.06 ± 0.55} \\
SL-GAT & \highlight{lightgray}{48.11 ± 2.18} & \highlight{lightgray}{35.75 ± 2.03} & \highlight{darkgray}{53.50 ± 0.42} \\
\midrule
Baseline-Tran & \highlight{darkgray}{68.43 ± 2.22} & \highlight{lightgray}{59.68 ± 1.89} & 45.87 ± 0.54 \\
Graphcon-Tran & \highlight{lightgray}{66.88 ± 2.16} & \highlight{lightgray}{59.68 ± 2.50} & 46.08 ± 0.50 \\
KuramotoGNN & 65.32 ± 2.29 & 56.44 ± 1.79 & \highlight{lightgray}{46.17 ± 0.84} \\
SL-Tran & 66.03 ± 2.13 & \highlight{darkgray}{61.82 ± 2.29} & \highlight{darkgray}{48.64 ± 0.63} \\
\bottomrule
\end{tabular}

%% file: Tables/graph_classification.tex
\begin{tabular}{c c c c}
\toprule
Model & MUTAG  & ENZYMES & PROTEINS  \\
\midrule
Baseline-GCN & \highlight{lightgray}{70.47 ± 9.79} & 28.03 ± 7.27 & 70.50 ± 3.36 \\
GraphCON-GCN & 69.21 ± 10.95 & \highlight{lightgray}{29.87 ± 8.38} & \highlight{lightgray}{70.62 ± 4.26} \\
Kuramoto-GCN & 59.05 ± 10.59 & 29.27 ± 5.87 & 68.39 ± 4.50 \\
SL-GCN & \highlight{darkgray}{77.47 ± 9.00} & \highlight{darkgray}{44.33 ± 9.83} & \highlight{darkgray}{71.20 ± 4.56} \\
\midrule
Baseline-GAT & 71.26 ± 11.87 & 44.57 ± 9.50 & \highlight{lightgray}{71.30 ± 4.42} \\
Graphcon-GAT & \highlight{lightgray}{71.89 ± 10.24} & \highlight{lightgray}{45.10 ± 11.23} & \highlight{darkgray}{71.98 ± 3.47} \\
Kuramoto-GAT & 71.58 ± 9.82 & 31.03 ± 8.79 & \highlight{lightgray}{71.30 ± 3.83} \\
SL-GAT & \highlight{darkgray}{77.79 ± 10.80} & \highlight{darkgray}{48.43 ± 9.14} & 70.89 ± 3.63\\
\midrule
Baseline-Tran & 61.26 ± 13.99 & 20.60 ± 4.87 & 69.77 ± 5.75\\
Graphcon-Tran & 64.00 ± 11.88 & 17.70 ± 5.04 & 67.98 ± 4.75 \\
Kuramoto-Tran & \highlight{lightgray}{68.53 ± 12.30} & \highlight{lightgray}{21.90 ± 5.54} & \highlight{lightgray}{71.12 ± 4.09}\\
SL-Tran & \highlight{darkgray}{78.84 ± 13.54} & \highlight{darkgray}{39.40 ± 6.87} & \highlight{darkgray}{71.70 ± 3.30} \\
\bottomrule
\end{tabular}

%% file: Tables/graph_regression_short.tex
\begin{tabular}{c c c c}
\toprule
Model & ESOL  & FreeSolv &  Lipophilicity \\
\midrule
Baseline-GCN & 0.827 ± 0.199 & 4.780 ± 1.934 & 0.848 ± 0.462\\
GraphCON-GCN & 0.956 ± 0.718 & 4.185 ± 1.468 & \highlight{lightgray}{0.498 ± 0.062}\\
Kuramoto-GCN & \highlight{lightgray}{0.757 ± 0.152} & \highlight{lightgray}{2.993 ± 1.311} & 0.562 ± 0.113\\
SL-GCN & \highlight{darkgray}{0.673 ± 0.166} & \highlight{darkgray}{1.949 ± 0.819} & \highlight{darkgray}{0.495 ± 0.056}\\
\midrule
Baseline-GAT & \highlight{darkgray}{0.565 ± 0.116} & 2.903 ± 1.926 & 0.684 ± 0.457\\
Graphcon-GAT & 0.614 ± 0.125 & 5.938 ± 3.986 & \highlight{lightgray}{0.542 ± 0.176}\\
Kuramoto-GAT & 0.648 ± 0.128 & \highlight{lightgray}{2.130 ± 1.020} & 0.579 ± 0.082\\
SL-GAT & \highlight{lightgray}{0.606 ± 0.139} & \highlight{darkgray}{2.014 ± 0.545} & \highlight{darkgray}{0.449 ± 0.069}\\
\midrule
Baseline-Tran & 1.887 ± 0.629 & 7.406 ± 2.901 & 1.205 ± 0.173 \\
Graphcon-Tran & \highlight{lightgray}{1.799 ± 0.619} & 10.677 ± 3.176 & 1.426 ± 0.082\\
Kuramoto-Tran & 1.943 ± 0.509 & \highlight{lightgray}{4.304 ± 1.484} & \highlight{lightgray}{0.879 ± 0.226}\\
SL-Tran & \highlight{darkgray}{1.187 ± 0.259} & \highlight{darkgray}{3.877 ± 1.783} & \highlight{darkgray}{0.695 ± 0.087}\\
\bottomrule 
\end{tabular}

%% file: Tables/t-test_homophilic.tex
\begin{tabular}{l c c c c c c}
\toprule
Model & Cora & Citeseer & Pubmed & Texas & Wisconsin & Cornell \\
\midrule
SL-GCN & \textbf{2.67} & 0.46 & \textbf{4.81} & 1.61 & 0.44 & 0.57\\
SL-GAT & 1.39 & 1.30 & 0.62  & \textbf{2.06} & 1.15 & 0.14\\
SL-Tran & \textbf{2.84} & \textbf{3.52} & \textbf{1.73} & \textbf{2.86} & 0.32 & \textbf{5.17} \\
\bottomrule
\end{tabular}

%% file: Tables/t-test_homophilic2.tex
\begin{tabular}{l c c c c c c}
\toprule
Model & Chameleon & Squirrel & Amazon-rating & MUTAG  & ENZYMES & PROTEINS \\
\midrule
SL-GCN & 1.10 & \textbf{2.97} & \textbf{26.20}  & \textbf{3.72} & \textbf{7.92} & 0.66 \\
SL-GAT & 0.68 & 1.05 & \textbf{35.26} & \textbf{2.80} & 1.63 & -0.51 \\
SL-Tran & -2.80 & \textbf{7.21} & 23.52 & \textbf{3.99} & \textbf{14.02} & 0.78 \\
\bottomrule
\end{tabular}

%% file: Tables/t-test_homophilic3.tex
\begin{tabular}{l c c c}
\toprule
Model & ESOL  & FreeSolv &  Lipophilicity \\
\midrule
SLGCN & \textbf{4.20} & \textbf{9.41} & \textbf{3.76} \\
SLGAT & 0.30 & \textbf{3.14} & \textbf{8.58} \\
SLTran & \textbf{7.28} & \textbf{7.33} & \textbf{18.62} \\
\bottomrule
\end{tabular}

%% file: Tables/nlayer_corr_coeff.tex
\begin{tabular}{c c c c c}
\toprule
 & Baseline & GraphCON & Kuramoto & SLGNN \\
\midrule
\makecell{Correlation\\Coefficient} & 0.223 & 0.148 & 0.159 & 0.313 \\
\bottomrule
\end{tabular}

%% file: DRAFT/bibliography.bib
@article{arenas-diaz-kurths-moreno-zhou,
title = {Synchronization in complex networks},
author = {Arenas, Alex and Díaz-Guilera, Albert and Kurths, Jurgen and Moreno, Yamir and Zhou, Changsong},
journal = {Physics Reports},
Volume ={469},
Issue = {3},
year ={2008},
Pages = {93-153},
ISSN = {0370-1573},
doi = {https://doi.org/10.1016/j.physrep.2008.09.002.
(https://www.sciencedirect.com/science/article/pii/S0370157308003384)}
}

@inproceedings{chamberlainGRANDGraphNeural2021,
  title = {{{GRAND}}: {{Graph Neural Diffusion}}},
  shorttitle = {Grand},
  booktitle = {{{ICML}}},
  author = {Chamberlain, Benjamin Paul and Rowbottom, James and Gorinova, Maria and Webb, Stefan and Rossi, Emanuele and Bronstein, Michael M.},
  year = {2021},
  month = jun,
  eprint = {2106.10934},
  primaryclass = {cs, stat},
  urldate = {2024-10-07},
  archiveprefix = {arXiv},
  langid = {english}
}

@inproceedings{damianouGraphFoundationModels2024,
  title = {Towards {{Graph Foundation Models}} for {{Personalization}}},
  booktitle = {{{WWW}}},
  author = {Damianou, Andreas and Fabbri, Francesco and Gigioli, Paul and De Nadai, Marco and Wang, Alice and Palumbo, Enrico and Lalmas, Mounia},
  year = {2024},
  month = mar,
  eprint = {2403.07478},
  primaryclass = {cs},
  doi = {10.48550/arXiv.2403.07478},
  urldate = {2024-06-18},
  archiveprefix = {arXiv},
  langid = {english}
}

@article{deco,
  author = {Ponce-Alvarez, A and Deco, G},
  title = {The Hopf whole-brain model and its linear approximation},
  journal = {Sci Rep},
  volume = {14},
  pages = {2615},
  year = {2024},
  doi = {10.1038/s41598-024-53105-0}
}

@inproceedings{fanGraphNeuralNetworks2019,
  title = {Graph Neural Networks for Social Recommendation},
  booktitle = {{{WWW}}},
  author = {Fan, Wenqi and Ma, Yao and Li, Qing and He, Yuan and Zhao, Eric and Tang, Jiliang and Yin, Dawei},
  year = {2019},
  month = feb,
  eprint = {1902.07243},
  primaryclass = {cs},
  doi = {10.1145/3308558.3313488},
  urldate = {2024-07-27},
  archiveprefix = {arXiv},
  langid = {english}
}

@article{guoTamingOversmoothingRepresentation2023,
  title = {Taming Over-Smoothing Representation on Heterophilic Graphs},
  author = {Guo, Kai and Cao, Xiaofeng and Liu, Zhining and Chang, Yi},
  year = {2023},
  month = nov,
  journal = {Information Sciences},
  volume = {647},
  pages = {119463},
  issn = {00200255},
  doi = {10.1016/j.ins.2023.119463},
  urldate = {2025-05-09},
  langid = {english}
}

@inproceedings{kipfSemisupervisedClassificationGraph2016,
  title = {Semi-Supervised Classification with Graph Convolutional Networks},
  booktitle = {{{ICLR}}},
  author = {Kipf, Thomas N. and Welling, Max},
  year = {2016},
  month = sep,
  eprint = {1609.02907},
  urldate = {2024-11-15},
  archiveprefix = {arXiv},
  langid = {english}
}

@article{Kuramoto:1975ebm,
	author = {Kuramoto, Yoshiki},
	doi = {10.1007/BFb0013365},
	editor = {Araki, Huzihiro},
	journal = {Lect. Notes Phys.},
	pages = {420--422},
	title = {{Self-entrainment of a population of coupled non-linear oscillators}},
	volume = {39},
	year = {1975}}

@inproceedings{lanthalerNeuralOscillatorsAre2023,
  title = {Neural {{Oscillators}} Are {{Universal}}},
  booktitle = {{{NeurIPS}}},
  author = {Lanthaler, Samuel and Rusch, T. Konstantin and Mishra, Siddhartha},
  year = {2023},
  month = may,
  eprint = {2305.08753},
  primaryclass = {cs},
  doi = {10.48550/arXiv.2305.08753},
  urldate = {2025-03-05},
  archiveprefix = {arXiv}
}

@article{MPRT-SL,
    author = {Millán, Ana P and Poyato, David and Reynolds, David N and Tudisco, Francesco},
    title = {Synchronization of coupled Stuart-Landau oscillators: How heterogeneity can facilitate synchronization},
    journal = {arXiv:2510.05243},
    year = {2025}
}

@article{miyatoArtificialKuramotoOscillatory2024,
  title = {Artificial {{Kuramoto Oscillatory Neurons}}},
  author = {Miyato, Takeru and L{\"o}we, Sindy and Geiger, Andreas and Welling, Max},
  year = {2024},
  month = oct,
  number = {arXiv:2410.13821},
  eprint = {2410.13821},
  publisher = {arXiv},
  doi = {10.48550/arXiv.2410.13821},
  urldate = {2024-10-21},
  archiveprefix = {arXiv},
  langid = {english}
}

@article{RevModPhys.90.031001,
	author = {Mu\~noz, Miguel A.},
	doi = {10.1103/RevModPhys.90.031001},
	issue = {3},
	journal = {Rev. Mod. Phys.},
	month = {Jul},
	numpages = {30},
	pages = {031001},
	publisher = {American Physical Society},
	title = {Colloquium: Criticality and dynamical scaling in living systems},
	url = {https://link.aps.org/doi/10.1103/RevModPhys.90.031001},
	volume = {90},
	year = {2018}}

@article{nguyenCoupledOscillatorsGraph2023,
  title = {From {{Coupled Oscillators}} to {{Graph Neural Networks}}: {{Reducing Over-smoothing}} via a {{Kuramoto Model-based Approach}}},
  shorttitle = {From Coupled Oscillators to Graph Neural Networks},
  author = {Nguyen, Tuan and Honda, Hirotada and Sano, Takashi and Nguyen, Vinh and Nakamura, Shugo and Nguyen, Tan M.},
  year = {2023},
  month = nov,
  journal = {International Conference on Artificial Intelligence and Statistics},
  eprint = {2311.03260},
  primaryclass = {cs},
  pages = {2710--2718},
  doi = {10.48550/arXiv.2311.03260},
  urldate = {2025-01-17},
  archiveprefix = {arXiv},
  langid = {english}
}

@inproceedings{nguyenRevisitingOversmoothingOversquashing2022,
  title = {Revisiting Over-Smoothing and over-Squashing Using Ollivier's Ricci Curvature},
  booktitle = {{{ICML}}},
  author = {Nguyen, Khang and Nong, Hieu and Nguyen, Vinh and Ho, Nhat and Osher, Stanley and Nguyen, Tan},
  year = {2022},
  month = nov,
  eprint = {2211.15779},
  primaryclass = {cs, stat},
  doi = {10.48550/arXiv.2211.15779},
  urldate = {2024-04-23},
  archiveprefix = {arXiv},
  langid = {english}
}

@article{ntRevisitingGraphNeural2019,
  title = {Revisiting Graph Neural Networks: All We Have Is Low-Pass Filters},
  shorttitle = {Revisiting Graph Neural Networks},
  author = {Nt, Hoang and Maehara, Takanori},
  year = {2019},
  month = may,
  number = {arXiv:1905.9550},
  eprint = {1905.9550},
  primaryclass = {stat},
  publisher = {arXiv},
  doi = {10.48550/arXiv.1905.09550},
  urldate = {2025-01-25},
  archiveprefix = {arXiv},
  langid = {english}
}

@inproceedings{oonoGraphNeuralNetworks2019,
  title = {Graph {{Neural Networks Exponentially Lose Expressive Power}} for {{Node Classification}}},
  booktitle = {{{ICLR}}},
  author = {Oono, Kenta and Suzuki, Taiji},
  year = {2019},
  month = may,
  eprint = {1905.10947},
  primaryclass = {cs, stat},
  urldate = {2024-07-03},
  archiveprefix = {arXiv},
  langid = {english}
}

@article{panteleyStabilityRobustnessStuartlandau2015,
  title = {On the Stability and Robustness of Stuart-Landau Oscillators},
  author = {Panteley, Elena and Loria, Antonio and Ati, Ali El},
  year = {2015},
  journal = {IFAC-PapersOnLine},
  series = {1st {{IFAC Conference onModelling}}, {{Identification andControl}} of {{Nonlinear SystemsMICNON}} 2015},
  volume = {48},
  number = {11},
  pages = {645--650},
  issn = {24058963},
  doi = {10.1016/j.ifacol.2015.09.260},
  urldate = {2025-01-01},
  copyright = {https://www.elsevier.com/tdm/userlicense/1.0/},
  langid = {english}
}

@inproceedings{peiGeomGCNGeometricGraph2020,
  title = {Geom-{{GCN}}: {{Geometric Graph Convolutional Networks}}},
  shorttitle = {Geom-{{GCN}}},
  booktitle = {{{ICLR}}},
  author = {Pei, Hongbin and Wei, Bingzhe and Chang, Kevin Chen-Chuan and Lei, Yu and Yang, Bo},
  year = {2020},
  month = feb,
  eprint = {2002.05287},
  primaryclass = {cs},
  doi = {10.48550/arXiv.2002.05287},
  urldate = {2025-03-29},
  archiveprefix = {arXiv}
}

@inproceedings{pengSVDGCNSimplifiedGraph2022,
  title = {{{SVD-GCN}}: A Simplified Graph Convolution Paradigm for Recommendation},
  shorttitle = {Svd-Gcn},
  booktitle = {{{CIKM}}},
  author = {Peng, Shaowen and Sugiyama, Kazunari and Mine, Tsunenori},
  year = {2022},
  month = aug,
  eprint = {2208.12689},
  primaryclass = {cs},
  doi = {10.1145/3511808.3557462},
  urldate = {2024-07-01},
  archiveprefix = {arXiv},
  langid = {english}
}

@inproceedings{platonovCriticalLookEvaluation2023,
  title = {A Critical Look at the Evaluation of {{GNNs}} under Heterophily: Are We Really Making Progress?},
  shorttitle = {A Critical Look at the Evaluation of {{GNNs}} under Heterophily},
  booktitle = {{{ICLR}}},
  author = {Platonov, Oleg and Kuznedelev, Denis and Diskin, Michael and Babenko, Artem and Prokhorenkova, Liudmila},
  year = {2023},
  month = feb,
  eprint = {2302.11640},
  primaryclass = {cs},
  doi = {10.48550/arXiv.2302.11640},
  urldate = {2025-05-09},
  archiveprefix = {arXiv}
}

@book{Rao,
	author = {Rao, S.S.},
	date = {2004},
	publisher = {Pearson Prentice Hall},
	title = {Mechanical Vibrations},
	year = {2004}}

@article{Reynolds_2025,
	author = {Reynolds, David N and Tudisco, Francesco},
	doi = {10.1088/1361-6544/add3ae},
	journal = {Nonlinearity},
	month = {may},
	number = {5},
	pages = {055027},
	publisher = {IOP Publishing},
	title = {Unique Nash equilibrium of a nonlinear model of opinion dynamics on networks with friction-inspired stubbornness},
	url = {https://doi.org/10.1088/1361-6544/add3ae},
	volume = {38},
	year = {2025}}

@inproceedings{rothRankCollapseCauses2023,
  title = {Rank Collapse Causes Over-Smoothing and over-Correlation in Graph Neural Networks},
  booktitle = {{{LoG}}},
  author = {Roth, Andreas and Liebig, Thomas},
  year = {2023},
  issn = {2640-3498},
  urldate = {2024-11-11},
  langid = {english}
}

@inproceedings{ruschCoupledOscillatoryRecurrent2020,
  title = {Coupled {{Oscillatory Recurrent Neural Network}} ({{coRNN}}): {{An}} Accurate and (Gradient) Stable Architecture for Learning Long Time Dependencies},
  shorttitle = {Coupled {{Oscillatory Recurrent Neural Network}} ({{coRNN}})},
  booktitle = {{{ICLR}}},
  author = {Rusch, T. Konstantin and Mishra, Siddhartha},
  year = {2020},
  month = oct,
  eprint = {2010.00951},
  primaryclass = {cs},
  doi = {10.48550/arXiv.2010.00951},
  urldate = {2025-03-04},
  archiveprefix = {arXiv}
}

@inproceedings{ruschGraphCoupledOscillatorNetworks2022,
  title = {Graph-{{Coupled Oscillator Networks}}},
  booktitle = {{{ICML}}},
  author = {Rusch, T. Konstantin and Chamberlain, Benjamin P. and Rowbottom, James and Mishra, Siddhartha and Bronstein, Michael M.},
  year = {2022},
  month = feb,
  eprint = {2202.02296},
  primaryclass = {cs, math, stat},
  urldate = {2024-10-07},
  archiveprefix = {arXiv},
  langid = {english}
}

@article{ruschSurveyOversmoothingGraph2023,
  title = {A Survey on Oversmoothing in Graph Neural Networks},
  author = {Rusch, T. Konstantin and Bronstein, Michael M. and Mishra, Siddhartha},
  year = {2023},
  month = mar,
  number = {arXiv:2303.10993},
  eprint = {2303.10993},
  primaryclass = {cs},
  publisher = {arXiv},
  doi = {10.48550/arXiv.2303.10993},
  urldate = {2024-04-23},
  archiveprefix = {arXiv},
  langid = {english}
}

@inproceedings{ruschUnICORNNRecurrentModel2021,
  title = {{{UnICORNN}}: {{A}} Recurrent Model for Learning Very Long Time Dependencies},
  shorttitle = {{{UnICORNN}}},
  booktitle = {{{ICML}}},
  author = {Rusch, T. Konstantin and Mishra, Siddhartha},
  year = {2021},
  month = mar,
  eprint = {2103.05487},
  primaryclass = {cs},
  doi = {10.48550/arXiv.2103.05487},
  urldate = {2025-03-04},
  archiveprefix = {arXiv}
}

@article{Sclosa_2024,
	author = {Sclosa, Davide},
	doi = {10.1088/1361-6544/ad694a},
	journal = {Nonlinearity},
	month = {aug},
	number = {9},
	pages = {095021},
	publisher = {IOP Publishing},
	title = {Completely degenerate equilibria of the Kuramoto model on networks},
	url = {https://doi.org/10.1088/1361-6544/ad694a},
	volume = {37},
	year = {2024}}

@article{doi:10.1137/23M155400X,
	author = {Sclosa, Davide},
	doi = {10.1137/23M155400X},
	eprint = {https://doi.org/10.1137/23M155400X},
	journal = {SIAM Journal on Applied Dynamical Systems},
	number = {4},
	pages = {3267-3283},
	title = {Kuramoto Networks with Infinitely Many Stable Equilibria},
	url = {https://doi.org/10.1137/23M155400X},
	volume = {22},
	year = {2023}}

@article{shchurPitfallsGraphNeural2018,
  title = {Pitfalls of Graph Neural Network Evaluation},
  author = {Shchur, Oleksandr and Mumme, Maximilian and Bojchevski, Aleksandar and G{\"u}nnemann, Stephan},
  year = {2018},
  month = nov,
  number = {arXiv:1811.5868},
  eprint = {1811.5868},
  primaryclass = {cs},
  publisher = {arXiv},
  doi = {10.48550/arXiv.1811.05868},
  urldate = {2024-12-03},
  archiveprefix = {arXiv},
  langid = {english}
}

@article{Taylor_2012,
	author = {Taylor, Richard},
	doi = {10.1088/1751-8113/45/5/055102},
	journal = {Journal of Physics A: Mathematical and Theoretical},
	month = {jan},
	number = {5},
	pages = {055102},
	publisher = {IOP Publishing},
	title = {There is no non-zero stable fixed point for dense networks in the homogeneous Kuramoto model},
	url = {https://doi.org/10.1088/1751-8113/45/5/055102},
	volume = {45},
	year = {2012}}

@article{thorpeGrandGraphNeural2022,
  title = {Grand++: Graph Neural Diffusion with a Source Term},
  author = {Thorpe, Matthew and Xia, Hedi and Nguyen, Tan and Strohmer, Thomas and Bertozzi, Andrea L and Osher, Stanley J},
  year = {2022},
  langid = {english}
}

@inproceedings{velickovicGraphAttentionNetworks2017,
  title = {Graph Attention Networks},
  booktitle = {{{ICLR}}},
  author = {Veli{\v c}kovi{\'c}, Petar and Cucurull, Guillem and Casanova, Arantxa and Romero, Adriana and Li{\`o}, Pietro and Bengio, Yoshua},
  year = {2017},
  month = oct,
  eprint = {1710.10903},
  primaryclass = {cs, stat},
  doi = {10.17863/CAM.48429},
  urldate = {2024-07-25},
  archiveprefix = {arXiv},
  langid = {english}
}

@article{10.1063/1.2165594,
	author = {Wiley, Daniel A. and Strogatz, Steven H. and Girvan, Michelle},
	doi = {10.1063/1.2165594},
	issn = {1054-1500},
	journal = {Chaos: An Interdisciplinary Journal of Nonlinear Science},
	month = {03},
	number = {1},
	pages = {015103},
	title = {The size of the sync basin},
	url = {https://doi.org/10.1063/1.2165594},
	volume = {16},
	year = {2006}}

@inproceedings{xhonneuxContinuousGraphNeural2019,
  title = {Continuous {{Graph Neural Networks}}},
  booktitle = {{{ICML}}},
  author = {Xhonneux, Louis-Pascal A. C. and Qu, Meng and Tang, Jian},
  year = {2019},
  month = dec,
  eprint = {1912.00967},
  primaryclass = {cs},
  doi = {10.48550/arXiv.1912.00967},
  urldate = {2025-03-02},
  archiveprefix = {arXiv}
}

@InProceedings{pmlr-v70-gilmer17a,
  title = 	 {Neural Message Passing for Quantum Chemistry},
  author =       {Justin Gilmer and Samuel S. Schoenholz and Patrick F. Riley and Oriol Vinyals and George E. Dahl},
  booktitle = 	 {Proceedings of the 34th International Conference on Machine Learning},
  pages = 	 {1263--1272},
  year = 	 {2017},
  editor = 	 {Precup, Doina and Teh, Yee Whye},
  volume = 	 {70},
  series = 	 {Proceedings of Machine Learning Research},
  month = 	 {06--11 Aug},
  publisher =    {PMLR},
  url = 	 {https://proceedings.mlr.press/v70/gilmer17a.html}
}

@article{zhang2025rethinking,
  title={Rethinking Oversmoothing in Graph Neural Networks: A Rank-Based Perspective},
  author={Zhang, Kaicheng and Deidda, Piero and Higham, Desmond and Tudisco, Francesco},
  journal={arXiv preprint arXiv:2502.04591},
  year={2025}
}

@article{
giovanni2024how,
title={How does over-squashing affect the power of {GNN}s?},
author={Francesco Di Giovanni and T. Konstantin Rusch and Michael Bronstein and Andreea Deac and Marc Lackenby and Siddhartha Mishra and Petar Veli{\v{c}}kovi{\'c}},
journal={Transactions on Machine Learning Research},
issn={2835-8856},
year={2024},
url={https://openreview.net/forum?id=KJRoQvRWNs},
note={}
}

@article{poli2019graph,
  title={Graph neural ordinary differential equations},
  author={Poli, Michael and Massaroli, Stefano and Park, Junyoung and Yamashita, Atsushi and Asama, Hajime and Park, Jinkyoo},
  journal={arXiv preprint arXiv:1911.07532},
  year={2019}
}

@inproceedings{Eliasof,
 author = {Eliasof, Moshe and Haber, Eldad and Treister, Eran},
 booktitle = {Advances in Neural Information Processing Systems},
 editor = {M. Ranzato and A. Beygelzimer and Y. Dauphin and P.S. Liang and J. Wortman Vaughan},
 pages = {3836--3849},
 publisher = {Curran Associates, Inc.},
 title = {PDE-GCN: Novel Architectures for Graph Neural Networks Motivated by Partial Differential Equations},
 url = {https://proceedings.neurips.cc/paper_files/paper/2021/file/1f9f9d8ff75205aa73ec83e543d8b571-Paper.pdf},
 volume = {34},
 year = {2021}
}

@article{GraphNeuralReactionDiffusionModels_Eliasof,
author = {Eliasof, Moshe and Haber, Eldad and Treister, Eran},
title = {Graph Neural Reaction Diffusion Models},
journal = {SIAM Journal on Scientific Computing},
volume = {46},
number = {4},
pages = {C399-C420},
year = {2024},
doi = {10.1137/23M1576700},

URL = { 
    
        https://doi.org/10.1137/23M1576700
    
    

},
eprint = { 
    
        https://doi.org/10.1137/23M1576700
    
    

}
}

@InProceedings{GREAD_Choi,
  title = 	 {{GREAD}: Graph Neural Reaction-Diffusion Networks},
  author =       {Choi, Jeongwhan and Hong, Seoyoung and Park, Noseong and Cho, Sung-Bae},
  booktitle = 	 {Proceedings of the 40th International Conference on Machine Learning},
  pages = 	 {5722--5747},
  year = 	 {2023},
  editor = 	 {Krause, Andreas and Brunskill, Emma and Cho, Kyunghyun and Engelhardt, Barbara and Sabato, Sivan and Scarlett, Jonathan},
  volume = 	 {202},
  series = 	 {Proceedings of Machine Learning Research},
  month = 	 {23--29 Jul},
  publisher =    {PMLR},
  url = 	 {https://proceedings.mlr.press/v202/choi23a.html}
}

@inproceedings{eliasof2024featuretransportation,
  title={Feature transportation improves graph neural networks},
  author={Eliasof, Moshe and Haber, Eldad and Treister, Eran},
  booktitle={Proceedings of the AAAI conference on artificial intelligence},
  volume={38},
  number={11},
  pages={11874--11882},
  year={2024}
}

@article{mccallumAutomatingConstructionInternet2000,
  title = {Automating the Construction of Internet Portals with Machine Learning},
  author = {McCallum, Andrew Kachites and Nigam, Kamal and Rennie, Jason and Seymore, Kristie},
  date = {2000-07-01},
  journaltitle = {Information Retrieval},
  shortjournal = {Inf. Retr.},
  volume = {3},
  number = {2},
  pages = {127--163},
  issn = {1573-7659},
  doi = {10.1023/A:1009953814988},
  url = {https://doi.org/10.1023/A:1009953814988},
  urldate = {2025-10-02},
  langid = {english}
}

@inproceedings{namataQuerydrivenActiveSurveying2012,
  title = {Query-Driven {{Active Surveying}} for {{Collective Classification}}},
  author = {Namata, Galileo and London, Ben and Getoor, L. and Huang, Bert},
  date = {2012},
  url = {https://www.semanticscholar.org/paper/efac04450c531b3769451a886ed9a42fce4754d9},
  urldate = {2025-10-02},
  eventtitle = {10th {{International Workshop}} on {{Mining}} and {{Learning}} with {{Graphs}}},
  langid = {english}
}

@article{senCollectiveClassificationNetwork2008,
  title = {Collective Classification in Network Data},
  author = {Sen, Prithviraj and Namata, Galileo and Bilgic, Mustafa and Getoor, Lise and Galligher, Brian and Eliassi-Rad, Tina},
  date = {2008-09-06},
  journaltitle = {AI Magazine},
  shortjournal = {AI Mag.},
  volume = {29},
  number = {3},
  pages = {93--93},
  issn = {2371-9621},
  doi = {10.1609/aimag.v29i3.2157},
  url = {https://ojs.aaai.org/aimagazine/index.php/aimagazine/article/view/2157},
  urldate = {2025-10-02},
  langid = {english}
}

@article{rozemberczkiMultiscaleAttributedNode2019,
  title = {Multi-Scale {{Attributed Node Embedding}}},
  author = {Rozemberczki, Benedek and Allen, Carl and Sarkar, Rik},
  date = {2019-09-25},
  journaltitle = {Journal of Complex Networks},
  volume = {9},
  number = {2},
  eprint = {1909.13021},
  eprinttype = {arXiv},
  eprintclass = {cs},
  pages = {cnab014},
  doi = {10.1093/comnet/cnab014},
  url = {https://www.semanticscholar.org/paper/135334ea7fdef8eef0367e862797cac7dcd232a4},
  urldate = {2025-10-02},
  langid = {english}
}

@inproceedings{moritzRayDistributedFramework2017,
  title = {Ray: {{A Distributed Framework}} for {{Emerging AI Applications}}},
  shorttitle = {Ray},
  booktitle = {{{USENIX Symposium}} on {{Operating Systems Design}} and {{Implementation}}},
  author = {Moritz, Philipp and Nishihara, Robert and Wang, Stephanie and Tumanov, Alexey and Liaw, Richard and Liang, Eric and Elibol, Melih and Yang, Zongheng and Paul, William and Jordan, Michael I. and Stoica, Ion},
  date = {2017-12-16},
  volume = {abs/1712.05889},
  eprint = {1712.05889},
  eprinttype = {arXiv},
  eprintclass = {cs},
  publisher = {arXiv},
  doi = {10.48550/arXiv.1712.05889},
  url = {https://www.semanticscholar.org/paper/f83a207712fd4cf41aded79e9e6c4345ba879128},
  urldate = {2025-10-02},
  langid = {english}
}

@inproceedings{akibaOptunaNextgenerationHyperparameter2019,
  title = {Optuna: {{A Next-generation Hyperparameter Optimization Framework}}},
  shorttitle = {Optuna},
  booktitle = {Knowledge {{Discovery}} and {{Data Mining}}},
  author = {Akiba, Takuya and Sano, Shotaro and Yanase, Toshihiko and Ohta, Takeru and Koyama, Masanori},
  date = {2019-07-25},
  eprint = {1907.10902},
  eprinttype = {arXiv},
  eprintclass = {cs},
  publisher = {arXiv},
  doi = {10.1145/3292500.3330701},
  url = {https://www.semanticscholar.org/paper/4cdf2fad22afc865999747336c7399fe422e6e8e},
  urldate = {2025-10-02},
  langid = {english}
}

@article{ivanovUnderstandingIsomorphismBias2019,
  title = {Understanding {{Isomorphism Bias}} in {{Graph Data Sets}}},
  author = {Ivanov, Sergei and Sviridov, Sergei and Burnaev, Evgeny},
  date = {2019-09-25},
  journaltitle = {arXiv.org},
  volume = {abs/1910.12091},
  eprint = {1910.12091},
  eprinttype = {arXiv},
  eprintclass = {cs},
  issn = {2331-8422},
  doi = {10.48550/arXiv.1910.12091},
  url = {https://www.semanticscholar.org/paper/1ec06d7591ceb91343648cc151546862447f1c9d},
  urldate = {2025-10-02},
  langid = {english}
}

@article{wuMoleculeNetBenchmarkMolecular2017,
  title = {{{MoleculeNet}}: A Benchmark for Molecular Machine Learning},
  shorttitle = {{{MoleculeNet}}},
  author = {Wu, Zhenqin and Ramsundar, Bharath and Feinberg, Evan N. and Gomes, Joseph and Geniesse, Caleb and Pappu, Aneesh S. and Leswing, Karl and Pande, Vijay},
  date = {2017-03-02},
  journaltitle = {Chemical Science},
  volume = {9},
  eprint = {1703.00564},
  eprinttype = {arXiv},
  eprintclass = {cs},
  pages = {513--530},
  issn = {2041-6520},
  doi = {10.1039/c7sc02664a},
  url = {https://www.semanticscholar.org/paper/d0ab11de3077490c80a08abd0fb8827bac84c454},
  urldate = {2025-10-02},
  langid = {english}
}

@article{huOpenGraphBenchmark2020,
  title = {Open {{Graph Benchmark}}: {{Datasets}} for {{Machine Learning}} on {{Graphs}}},
  shorttitle = {Open Graph Benchmark},
  author = {Hu, Weihua and Fey, Matthias and Zitnik, Marinka and Dong, Yuxiao and Ren, Hongyu and Liu, Bowen and Catasta, Michele and Leskovec, Jure},
  date = {2020-05-02},
  journaltitle = {Neural Information Processing Systems},
  volume = {abs/2005.00687},
  eprint = {2005.00687},
  eprinttype = {arXiv},
  eprintclass = {cs},
  doi = {10.48550/arXiv.2005.00687},
  url = {https://www.semanticscholar.org/paper/597bd2e45427563cdf025e53a3239006aa364cfc},
  urldate = {2025-10-02},
  langid = {english}
}

@article{ruschOscillatoryStateSpaceModels2024,
  title = {Oscillatory {{State-Space Models}}},
  author = {Rusch, T. Konstantin and Rus, Daniela},
  date = {2024-10-04},
  journaltitle = {International Conference on Learning Representations},
  volume = {abs/2410.03943},
  eprint = {2410.03943},
  eprinttype = {arXiv},
  eprintclass = {cs},
  doi = {10.48550/arXiv.2410.03943},
  url = {https://www.semanticscholar.org/paper/3bdeb86b4f6251a4c252f5e39a5db78620300bbc},
  urldate = {2025-08-22}
}

@article{fengGraphRandomNeural2020,
  title = {Graph {{Random Neural Networks}} for {{Semi-Supervised Learning}} on {{Graphs}}},
  author = {Feng, Wenzheng and Zhang, Jie and Dong, Yuxiao and Han, Yu and Luan, Huanbo and Xu, Qian and Yang, Qiang and Kharlamov, Evgeny and Tang, Jie},
  date = {2020-05-22},
  journaltitle = {Neural Information Processing Systems},
  eprint = {2005.11079},
  eprinttype = {arXiv},
  eprintclass = {cs},
  doi = {10.48550/arXiv.2005.11079},
  url = {https://www.semanticscholar.org/paper/7168f3bbd3b330af95dd698a86fe6ddcda5d6bbd},
  urldate = {2025-10-07},
  langid = {english}
}


%% file: sample-base.bib
@String{Computing = "Computing" }

@ArtifactSoftware{R,
    title = {R: A Language and Environment for Statistical Computing},
    author = {{R Core Team}},
    organization = {R Foundation for Statistical Computing},
    address = {Vienna, Austria},
    year = {2019},
    url = {https://www.R-project.org/},
}
